\documentclass[lettersize,journal]{IEEEtran}
\usepackage{amsmath,amsfonts}
\usepackage{algorithmic}
\usepackage{algorithm}
\usepackage{array}
\usepackage[caption=false,font=normalsize,labelfont=sf,textfont=sf]{subfig}
\usepackage{textcomp}
\usepackage{stfloats}
\usepackage{url}
\usepackage{verbatim}
\usepackage{graphicx}
\usepackage{cite}

\usepackage{algorithmic}
\usepackage{textcomp}
\usepackage{color, soul}
\usepackage{tabu}                     
\usepackage{multicol}                 
\usepackage{multirow}                
\usepackage{float}                    
\usepackage{makecell}                 
\usepackage{booktabs}                 
\usepackage{colortbl}
\usepackage{pifont} 

\usepackage{amssymb}
\usepackage{graphicx}
\usepackage{mathrsfs}
\usepackage{amsmath}
\usepackage{makecell}                 
\usepackage{graphicx}
\usepackage{caption} 
\usepackage{subcaption}
\usepackage{ulem}

\usepackage[utf8]{inputenc} 
\usepackage[T1]{fontenc}    
\usepackage{hyperref}       
\usepackage{url}            
\usepackage{booktabs}       
\usepackage{amsfonts}       
\usepackage{nicefrac}       
\usepackage{microtype}      
\usepackage{xcolor}         

\begin{document}

\title{EAR: Enhancing Uni-Modal Representations for Weakly Supervised Audio-Visual Video Parsing}

\author{Huilai Li, 
Xiaomeng Di, 
Ying Xing, 
Yonghao Dang, 
Yiming Wang, 
Jianqin Yin*, \IEEEmembership{Senior Member, IEEE}
\thanks{This work was supported by the National Key R\&D Program of China (2024YFC3015604), the Beijing Natural Science Foundation under Grant F2024203115, and the China Postdoctoral Science Foundation under Grant Number 2024M750255.}
\thanks{Huilai Li, Ying Xing, Yiming Wang and Jianqin Yin* are with the School of Intelligent Engineering and Automation, Beijing University of Posts and Telecommunications, Beijing 100876, China (e-mail: lihuilai@bupt.edu.cn; 	xingying@bupt.edu.cn; ymwang99@bupt.edu.cn; jqyin@bupt.edu.cn) (Corresponding author: Jianqin Yin).}
\thanks{Xiaomeng Di is with the State Grid Corporation of China, Beijing 100192, China (e-mail: dixmustb@163.com).}
\thanks{Yonghao Dang is with the School of Artificial Intelligence, Beijing University of Posts and Telecommunications, Beijing 100876, China (e-mail: dyh2018@bupt.edu.cn).}
}



\maketitle

\begin{abstract}
Weakly supervised Audio-Visual Video Parsing (AVVP) aims to recognize and temporally localize audio, visual, and audio-visual events in videos using only coarse-grained labels. Faced with the challenging task settings, existing research advances along two main paths: pre-training pseudo-label generators for fine-grained cross-modal semantic guidance, or refining AVVP model architectures to enhance audio-visual fusion. However, since audio and visual signals are typically unaligned, achieving accurate video parsing fundamentally relies on precise perception of uni-modal events. Yet these multi-modal focused strategies excessively emphasize multi-modal fusion while inadequately guiding and preserving uni-modal semantics, resulting in noisy pseudo-labels and sub-optimal video parsing performance. This paper proposes a novel framework that enhances uni-modal representations for both the pseudo-label generator and the AVVP model. Specifically, we introduce a similarity-based label migration approach to annotate pre-training data, thereby enabling the pseudo-label generator to better understand uni-modal events. We also employ a soft-constrained manner to refine modeling of uni-modal features in parallel with multi-modal fusion. These designs enable coordinated attention to both uni-modal and cross-modal representations, thus boosting the localization performance for events. Extensive experiments show that our method outperforms state-of-the-art methods in both pseudo-label and AVVP performance.
\end{abstract}


\section{Introduction}

\begin{figure*}[t]
	\centering
	\includegraphics[width=18cm,height=4.953cm]{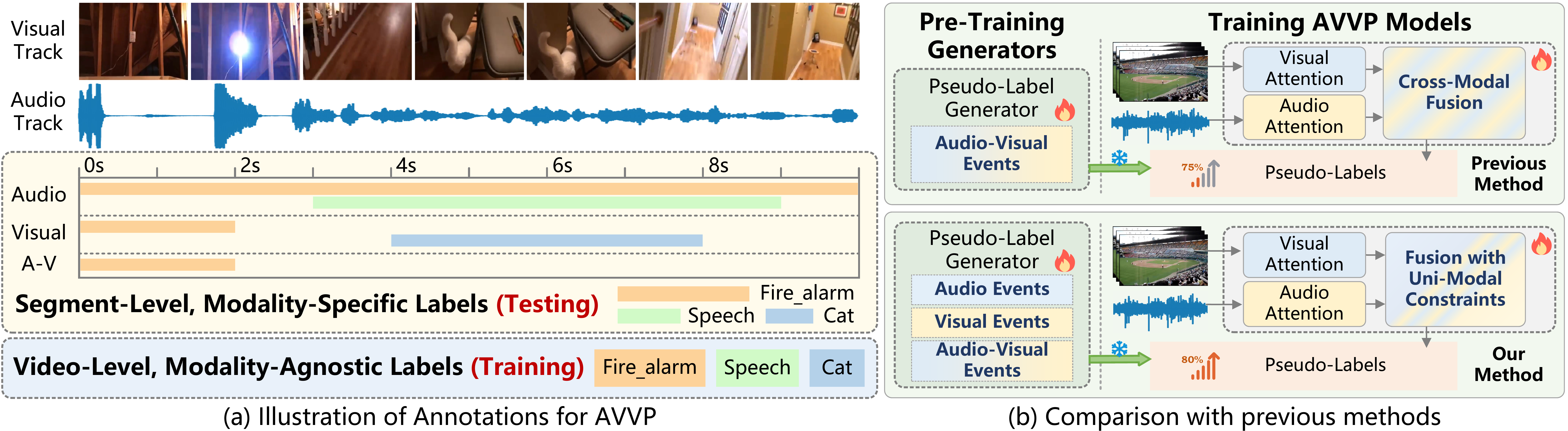}
	\caption{(a)Illustration of the AVVP. The testing data contains segment-level, modality-specific labels, while the training data only contains video-level, modality-agnostic labels. (b) Comparison of pseudo-label generator pre-training and AVVP model architecture between previous methods and the proposed method.}
	\label{introduction}
\end{figure*}

\IEEEPARstart{T}{he} Audio-Visual Video Parsing (AVVP) task is proposed to detect audio, visual, and audio-visual events in videos while simultaneously regressing their temporal boundaries. It facilitates robust scene understanding for machines by parsing complementary multi-modal cues even when events are partially observable or temporally misaligned. For example, as shown in Figure \ref{introduction} (a) from 2-10s, a system can perceive the audio event "fire\_alarm" despite the absence of its visual counterpart. This fundamental capability enables a wide range of applications such as sports analytics \cite{sadlier2005event, shih2017survey}, human-computer interaction \cite{fu2008real, sharma2021audio, chen2023question} and automated video editing \cite{liang2024language}. 

However, the annotation cost for AVVP is prohibitively expensive, as it requires precise temporal and modality-specific labeling. As a result, current training data only contains video-level, modality-agnostic labels, lacking detailed labels that specify the temporal boundaries and modality categories of events, as shown in Figure \ref{introduction} (a). To achieve accurate perception of both uni-modal and multi-modal events with only coarse-grained annotations, existing research primarily focuses on two directions: \emph{generating high-quality pseudo-labels} for finer cross-modal supervision \cite{fan2023revisit, zhou2024advancing, lai2023modality, lai2025uwav} or \emph{refining AVVP model architecture} for enhanced multi-modal representation \cite{tian2020unified, xu2024rethink, sardari2024coleaf}. Both strategies are driven by the need to establish reliable cross-modal correspondences, either by improving semantic guidance or strengthening multi-modal fusion. Nevertheless, this focus often comes at the expense of adequately learning and preserving the information of each modality, which carries independent yet crucial semantic cues for comprehensive understanding in AVVP. Therefore, in AVVP scenarios characterized by temporal misalignment between events, the uncoordinated modeling of uni-modal information in these methods may ultimately degrade detection accuracy for both uni-modal and multi-modal events.

Specifically, the strategy of \emph{generating high-quality pseudo-labels} lacks semantic guidance from individual modalities during the pre-training of the pseudo-label generator; the strategy of \emph{refining AVVP model architectures} lacks adequate modeling and preservation of information from uni-modal events. \textbf{(1) Pseudo-labeling approaches} \cite{fan2023revisit, zhou2024advancing, lai2023modality} typically leverage CLIP \cite{radford2021learning} and CLAP \cite{wu2023large} to generate segment-level pseudo-labels. Recent work \cite{lai2025uwav} further pre-trains generators with \emph{dense audio-visual event localization} data \cite{geng2023dense} to enhance their temporal modeling capability for inter-segment dynamic relationships. Despite enhancing overall performance, the absence of uni-modal supervision during pre-training inevitably leads to noisy pseudo-label generation. For audio-visual events, the pseudo-label generators can learn through direct supervision, whereas for uni-modal events, they can only infer from implicit inter-segment associations. This imbalanced focus causes the generators to prioritize multi-modal events, but its understanding of uni-modal events remains vague. \textbf{(2) Architecture refinement approaches} seek to enhance the video parsing performance by constructing sophisticated cross-modal fusion mechanism. A typical line of research \cite{tian2020unified, wang2025link, xie2025segment} employs hybrid attention networks to learn video semantics through joint intra-modal and cross-modal modeling. However, they emphasize multi-modal fusion but neglect the independent semantic modeling of each modality, leading to ambiguous attention to uni-modal events. Only a few studies \cite{xu2024rethink, sardari2024coleaf} argue that cross-modal learning in weakly supervised scenarios can impair the modeling of individual modalities, but their multi-branch fusion designs are computationally costly. In summary, balancing multi-modal integration while avoiding interference with uni-modal information is also a crucial issue in audio-visual video parsing. 

To address the insufficient uni-modal event guidance during the pre-training of the pseudo-label generator and inadequate modeling of uni-modal information in the AVVP training process, we propose an enhanced uni-modal representation (EAR) framework that generates accurate segment-level, modality-specific pseudo-labels and improves final video parsing performance. The key differences between previous methods and our method are summarized in Figure \ref{introduction} (b). \textbf{(1)} Specifically, we introduce a similarity-based uni-modal label migration approach to explicitly enhance the understanding of the pseudo-label generator for uni-modal events. Based on the principle that segments containing identical events tend to be similar, we utilize inter-segment similarity within each modality to assign the same labels to highly similar segments. For example, there is an audio-visual event "playing tennis" in 1-4s. If the visual information in 4-6s is similar to it, the visual event "playing tennis" can also be considered to exist in 4-6s. In this way, we transform the audio-visual event localization data into AVVP data (containing audio, visual, and audio-visual event labels) for the pre-training of the pseudo-label generator. Additionally, an asymmetric temporal modeling architecture is applied, enabling the model to focus more on the dynamic relationships between segments in uni-modal events. The proposed improvements enhance the event perception capability of the generator, thereby producing more accurate pseudo-labels that effectively support AVVP training. \textbf{(2)} We adopt a soft-constrained manner to guide the AVVP model in preserving uni-modal semantic information during multi-modal learning. First, we design an asymmetric audio(visual)-driven fusion that utilizes refined audio features to query original visual features, thereby mitigating the negative impact of visual semantic information on audio features and vice versa. Moreover, our model incorporates dependencies among intra-modal and inter-modal events (e.g., lightning is often accompanied by rain), thereby constraining the model to achieve coordinated attention to both uni-modal and multi-modal events at the semantic level.

By enhancing the semantic guidance and modeling of uni-modal information, our framework introduces improvements in both pseudo-label generation and model architecture. Experimental evaluations show that our method achieves average performance improvements of 2.4\%-4.6\% in pseudo-label generation and 0.9\%-4.2\% in video parsing over existing mainstream researches. Our contributions are as follows:
\begin{itemize}
\item We propose EAR, a novel and effective framework for audio-visual video parsing that enables enhanced uni-modal guidance and modeling.
\item We introduce a similarity-based label migration approach and a soft-constrained approach to enhance perception of uni-modal events, where the former provides richer semantic guidance for the pseudo-label generator and the latter refines the parsing model architecture.
\item By improving the AVVP pseudo-label generation process, we produce segment-level, modality-specific pseudo-labels that achieve the highest accuracy among existing methods and will be released publicly.
\item Extensive experiments demonstrate the effectiveness of EAR on the AVVP task, achieving superior overall performance compared to state-of-the-art methods.
\end{itemize}

\section{Related work}
\subsection{Audio-Visual Video Parsing}
Audio-Visual Video Parsing was first proposed by Tian \emph{et al.} \cite{tian2020unified}, enabling models to understand uni-modal and multi-modal events in videos via a weakly-supervised approach. According to the enhancement strategies, existing studies can be divided into two branches: 

One major branch focuses on enhancing the performance of models by leveraging fine-grained pseudo-labels. Specifically, MA \cite{wu2021exploring} and JoMoLD \cite{cheng2022joint} leverage comparative information across videos to refine video-level labels, while PPL \cite{rachavarapu2024weakly} develop a prototype-based module for segment-level label estimation. However, as the detailed information inherently is extracted from coarse-grained annotations, these methods still struggled to achieve generalized AVVP with limited training data. Subsequent studies \cite{fan2023revisit, zhou2024advancing, lai2023modality} advanced pseudo-label generation by incorporating external knowledge from pre-trained models. Recently, Lai \emph{et al.} \cite{lai2025uwav} further refined the generation of segment-level, modality-specific pseudo-labels by modeling inter-segment dynamic relations, thereby achieving more accurate AVVP performance. Nevertheless, as their pseudo-label generators \cite{lai2025uwav} are dependent on pre-training with audio-visual event data \cite{geng2023dense}, their perception of uni-modal events remains constrained by the coverage and quality of the training data. Our method can guide the generator with both uni-modal and multi-modal events, which consequently yields pseudo-labels with reduced noise on the target dataset.

Studies in the second branch are dedicated to designing efficient model architectures. Tian \emph{et al.} \cite{tian2020unified} introduced HAN, a framework designed for joint uni-modal and multi-modal temporal modeling, which has since been widely used in subsequent work \cite{chen2024cm, sun2024multi, wang2025mug, chen2025teacher, chen2025ten, li2024exploring}. Additionally, other studies further designed semantically-driven model architectures. For example, Zhao \emph{et al.} \cite{zhao2025text} and Chen \emph{et al.} \cite{chen2025temtg} introduced textual features to enhance the semantic modeling, while Zhao \emph{et al.} \cite{zhao2025multimodal} avoided interference among events by decoupling features of different event categories. However, existing methods focus on cross-modal information fusion, while rarely considering the interference between audio and video. Although a few studies \cite{gao2023collecting, xu2024rethink, sardari2024coleaf} paid attention to the isolated modeling of audio and video, most of them introduced redundant branches, leading to high computational costs. Moreover, excessive constraints in such designs may weaken cross-modal complementary information. In this paper, we adopt a soft-constrained fusion strategy, ensuring the model maintains coordinated attention to uni-modal and multi-modal events.

\subsection{More Video Event Localization}
\noindent \textbf{Audio-Visual Event Localization (AVE):}\; AVE was proposed by Tian \emph{et al.} \cite{tian2018audio} to localize events that are both audible and visible in videos. To facilitate cross-modal audio-visual alignment and fusion, some studies \cite{xu2020cross, rao2022dual, he2024cace, liu2025audio} introduced relation-aware architectures to extract complementary information from audio-visual events. Other researchers focused more on event-related semantic information, applying methods like global-local modeling \cite{wu2019dual} and contrastive learning \cite{jiang2023leveraging} to distinguish foreground from background in videos. Recent studies \cite{sun2025listen} even drawn inspiration from biological theories to design audio-visual fusion architectures, enhancing the ability of models to perceive events. However, the existing AVE dataset consists of specifically trimmed video clips, each containing only a single event, which severely limits its practical application.

\noindent \textbf{Dense Audio-Visual Event Localization (DAVE):}\; To better adapt to real-world scenarios, Geng \emph{et al.} \cite{geng2023dense} proposed dense audio-visual event localization, which localizes multiple audio-visual events from long untrimmed videos. Faced with this complex task setting, some studies performed audio-visual feature alignment through contrastive learning \cite{xing2024locality}, multi-scale encoding \cite{liu2025fasten}, or adaptive attention mechanisms \cite{ahmadian2025dense} to integrate multi-modal features, while others \cite{han2025temporal, li2025esg} reduced the semantic gap through cross-modal semantic alignment. Subsequent researches explored more challenging task settings. For instance, Zhou \emph{et al.} \cite{zhou2025clasp, zhou2025towards} investigated DAVE under weakly supervised and open-vocabulary conditions, while Geng \emph{et al.} \cite{geng2025longvale} and Tang \emph{et al.} \cite{tang2024avicuna} employed large language models for joint multi-task audio-visual understanding. 

\noindent \textbf{Dense Video Captioning (DVC):}\; Compared to traditional Video Captioning \cite{gao2017video, han2024action}, which can only generate a global description for the video, DVC, first introduced by Krishna \emph{et al.} \cite{krishna2017dense}, focuses on event information and provides fine-grained descriptions. To perform localization and captioning simultaneously, some studies \cite{zhou2018end, wang2021end} adopted end-to-end frameworks to address these tasks in parallel, while others \cite{zhou2024streaming} achieved real-time DVC through streaming caption generation. With the development of large language models, some studies \cite{yang2023vid2seq} leveraged their temporal understanding and text generation capabilities to yield more accurate descriptions. Moreover, recent research \cite{iashin2020multi, fang2025multimodal} further incorporated multi-modal information, such as audio and text, contributing to the alignment of DVC with real-world scenarios.

\begin{figure*}[t]
	\centering
	\includegraphics[width=18cm,height=9.2cm]{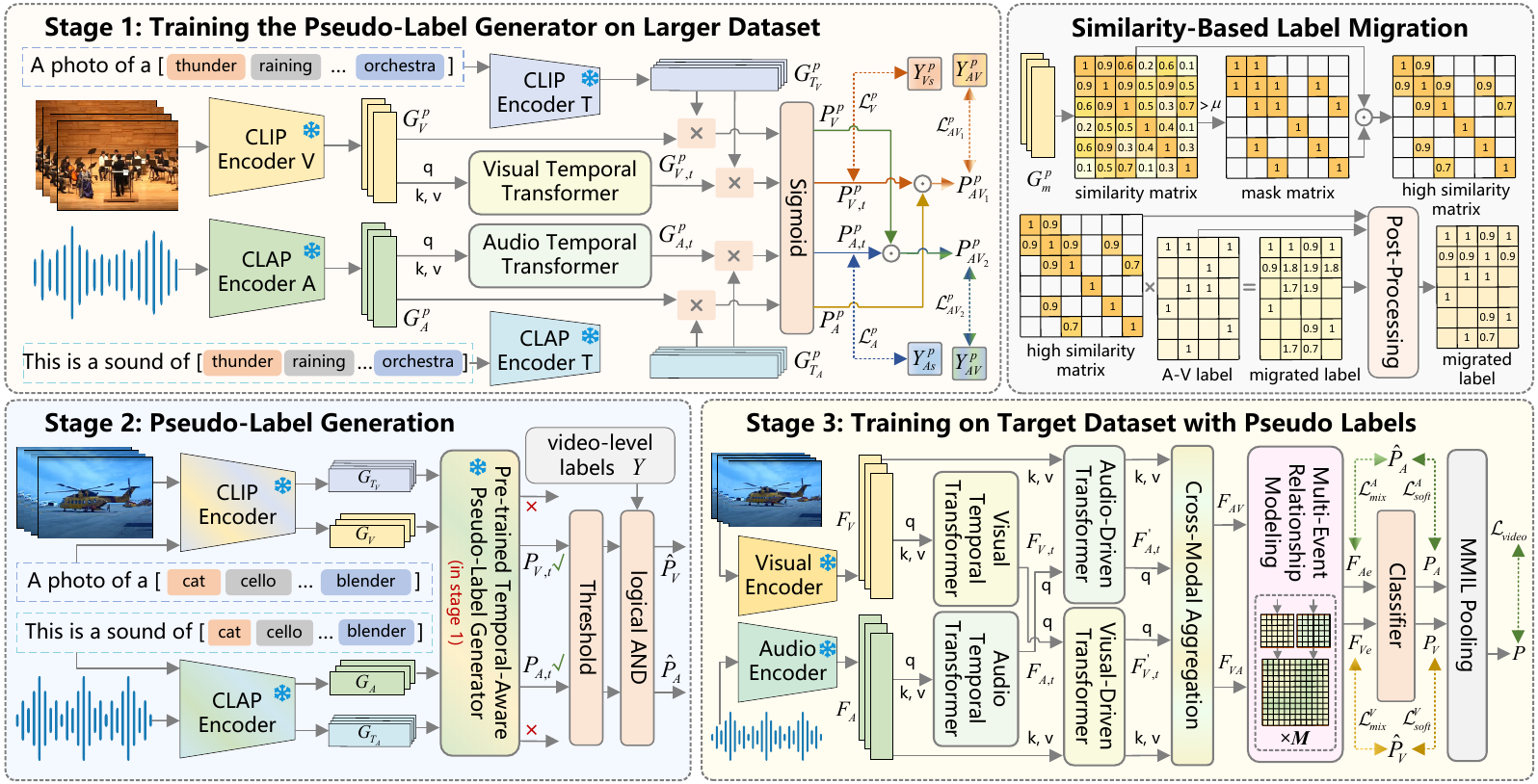}
	\caption{Overview of EAR. In stage 1, pre-training is conducted using a large-scale, dense audio-visual event localization dataset to equip the pseudo-label generator with temporal awareness for both uni-modal and multi-modal events. Note that the labels for audio and visual events are obtained via similarity-based label migration. Then, the pseudo-label generator is frozen to produce segment-level, modality-specific pseudo-labels for the target dataset. In stage 3, training is performed with the soft-constrained modeling and fine-grained label guidance for the audio-visual video parsing task.}
	\label{main_fig}
\end{figure*}

\section{Method}
In the AVVP task, each video is divided into $T$ non-overlapping audio and visual segments as $\{a_{t},v_{t}\}_{t=1}^{T}$. These segments are accompanied by a set of annotations $\{y_{t}^{a},y_{t}^{v},y_{t}^{av}\}_{t=1}^{T}\in\{0,1\}^{C}$, where $y_{t}^{a}$ and $y_{t}^{v}$ denote audio and visual events, $y_{t}^{av}$ denote audio-visual events that are the intersection of $y_{t}^{a}$ and $y_{t}^{v}$, and $C$ is the total number of event categories. Note that each segment may contain no events or multiple events. However, such fine-grained annotations are unavailable during training; only coarse-grained labels $Y\in\{0,1\}^{C}$ are provided. Since $y$ only indicate which events occur in the video, without localizing their temporal boundaries or specifying the modalities, existing models can only achieve event perception in a weakly-supervised manner.

\subsection{Overall Framework}
The proposed enhanced uni-modal representation framework can generate accurate pseudo-labels and allow the AVVP model to achieve greater attention to uni-modal events. Our EAR mainly consists of two components: \emph{Temporal-Aware Pseudo-Label Generation} and \emph{Semantic Modeling with Soft Constraints}. \textbf{(1)} As illustrated by Stage 1 and Stage 2 in Figure \ref{main_fig}, we first perform similarity-based uni-modal label migration on the large-scale DAVE dataset to pre-train the pseudo-label generator. Then, the temporal-aware generator is frozen and used to generate segment-level, modality-specific pseudo-labels on the AVVP dataset. The details are provided in Section \ref{pretrain}. \textbf{(2)} In Section \ref{softcon}, the pseudo-labels are used for weakly-supervised training on the target dataset with soft constraints.

\subsection{Temporal-Aware Pseudo-Label Generation}
\label{pretrain}

Previous methods \cite{lai2025uwav} pre-train the AVVP pseudo-label generator with the DAVE data \cite{geng2023dense}, which can capture temporal relationships between segments. However, due to the lack of uni-modal supervision, inference bias is introduced in the generator. In this subsection, we design a similarity-based label migration method and an asymmetric temporal modeling architecture to guide the pre-training with both uni-modal and multi-modal events, thereby generating more accurate pseudo-labels. 

\noindent \textbf{Similarity-Based Uni-Modal Label Migration:}
Visual or audio segments that are highly similar often contain the same events. Based on this principle, we can transfer labels from audio-visual events to uni-modal events, and use them for the pseudo-label generator pre-training. As shown in the upper-right corner of Figure \ref{main_fig}. We first extract uni-modal features $G_{m}^{p}\in\mathbb{R}^{T' \times D_{m}'}$ for $T'$ segments in a batch with the frozen CLIP \cite{radford2021learning} and CLAP \cite{wu2023large} model, where $m\in\{A,V\}$. Then, a cosine similarity matrix $S_{m}^{p}\in\mathbb{R}^{T' \times T'}$ between segments is derived to quantify their inter-segment associations. By setting a threshold $\mu_{m}$, we only retain the portions with high similarity, thereby filtering out irrelevant segment matches:
\begin{equation}
\hat{S}_{m}^{p}=M_{m}^{p} \odot S_{m}^{p}, 
M_{m}^{p}=\left\{
\begin{array}{ll}
1, & \text{if } S_{m}^{p} \geq \mu_{m} \\
0, & \text{otherwise}
\end{array}
\right.
\label{migration}
\end{equation}
where $M_{m}^{p}\in\mathbb{R}^{T' \times T'}$ represents the high similarity mask matrix, $\hat{S}_{m}^{p}\in\mathbb{R}^{T' \times T'}$ is the masked similarity matrix containing only high-similarity elements. By multiplying $\hat{S}_{m}^{p}$ with the audio-visual event label $Y_{AV}^{p}\in\mathbb{R}^{T' \times C'}$, uni-modal label migration can be achieved. That means by assigning identical labels to highly similar segments, the annotations of audio-visual events are extended to uni-modal event annotations:
\begin{equation}
Y_{ms}^{p}=\hat{S}_{m}^{p}\cdot Y_{AV}^{p}
\end{equation}
where $Y_{ms}^{p}\in\mathbb{R}^{T' \times C'}$ denotes the migrated uni-modal label, $C'$ is the number of event categories in DAVE dataset \cite{geng2023dense}. 

However, there may be multiple similar segments that contain the same audio-visual events, this leads to duplicate annotations in Eq. (\ref{migration}), resulting in some values greater than 1 in $Y_{ms}^{p}$. To avoid introducing additional noise and to meet the annotation requirements, we perform post-processing on the uni-modal labels $Y_{ms}^{p}$. For the duplicate annotations, we average them based on the number of times they are labeled, while for the annotations originally labeled as 1, we leave them unchanged (more details are shown in the supplementary material). Note that we do not use hard labels; instead, we adopt soft labels where similarity serves as the confidence score. Labels migrated based on high similarity have higher confidence, which further reduces the noise during the pre-training with uni-modal labels.

\noindent \textbf{Pre-Training of the Pseudo-Label Generator:}
After obtaining uni-modal and multi-modal labels, we can pre-train on the large-scale DAVE dataset \cite{geng2023dense} to enable the pseudo-label generator to capture dynamic temporal relationships between segments. To enhance the perception of the generator for uni-modal events, we design an asymmetric temporal modeling architecture. First, the visual and audio features $G_{m}^{p}$($m\in\{A,V\}$) are used for temporal modeling through single-modal transformer blocks to filter out noise and extract dynamic semantics between segments. Each transformer block consists of a multi-head self-attention (MSA) and a feed-forward network (FFN), as:
\begin{equation}
G_{m,t}^{p}=\text{FFN}[\text{MSA}(G_{m}^{p,q},G_{m}^{p,k},G_{m}^{p,v})]
\end{equation}
where $G_{m,t}^{p}\in\mathbb{R}^{T' \times D_{m}'}$ are temporal-aware uni-modal features. Meanwhile, we input all event categories from the pre-training dataset into fixed templates ("A photo of <event>" and "This is the sound of <event>"), and feed them into the frozen CLIP \cite{radford2021learning} and CLAP \cite{wu2023large} text encoders to extract the text label features $G_{T_{m}}^{p}\in\mathbb{R}^{C' \times D_{m}'}$. By aggregating the visual and audio features with the text features, the probabilities of uni-modal events is obtained as follows:
\begin{equation}
P_{m,t}^{p}=\text{Sigmoid}(G_{m,t}^{p}\cdot G_{T_{m}}^{p\top})
\end{equation}
\begin{equation}
P_{m}^{p}=\text{Sigmoid}(G_{m}^{p}\cdot G_{T_{m}}^{p\top})
\end{equation}
where $P_{m,t}^{p}\in\mathbb{R}^{T' \times C'}$ is the temporal-aware dynamic event probability, $P_{m}^{p}\in\mathbb{R}^{T' \times C'}$ is the static event probability based on single-frame inference. For the detection of audio-visual events, we avoid direct cross-modal modeling since it may interfere with uni-modal event detection, as in Figure \ref{introduction} (b). To this end, we propose an asymmetric architecture in which the dynamic visual/audio probability $P_{m,t}^{p}$ is combined with the static audio/visual probability $P_{m}^{p}$ to compute the audio-visual event probability:
\begin{equation}
P_{AV_{1}}^{p}=P_{V,t}^{p}\odot P_{A}^{p}
\end{equation}
\begin{equation}
P_{AV_{2}}^{p}=P_{A,t}^{p}\odot P_{V}^{p}
\end{equation}
where $P_{AV_{1}}^{p}, P_{AV_{2}}^{p}\in\mathbb{R}^{T' \times C'}$. Through the asymmetric architecture that simultaneously integrates dynamic and static event information, our model will rely on the optimization of temporal-aware uni-modal features $P_{m,t}^{p}$, thereby further enhancing the understanding of uni-modal events while ensuring the perception of multi-modal events. 

Finally, we use the ground truth of multi-modal events and the migrated uni-modal annotations to provide semantic guidance to the pseudo-label generator, which is pre-trained with the binary cross-entropy (BCE) loss:
\begin{multline}
\label{preloss}
\mathcal{L}_{pre} = \text{BCE}(P_{AV_{1}}^{p}, Y_{AV}^{p}) + \text{BCE}(P_{AV_{2}}^{p}, Y_{AV}^{p}) \\ 
+ \lambda_{A}\cdot\text{BCE}(P_{A,t}^{p}, Y_{As}^{p}) + \lambda_{V}\cdot\text{BCE}(P_{V,t}^{p}, Y_{Vs}^{p})
\end{multline}
where $\lambda_{A}$ and $\lambda_{V}$ are hyper-parameters used to mitigate the impact of noise introduced by migrated uni-modal labels.

\noindent \textbf{Pseudo-Label Generation on Target Dataset:}
The pseudo-label generator, pre-trained on large-scale DAVE data \cite{geng2023dense}, is now capable of accurate detection audio, visual, and audio-visual events. So it can be used to generate pseudo-labels on the target dataset of AVVP, as shown in Stage 2 of Figure \ref{main_fig}. We first use CLIP/CLAP \cite{radford2021learning, wu2023large} to extract uni-modal features $G_{m}\in\mathbb{R}^{T \times D_{m}^{g}}$ and the textual features $G_{T_{m}}\in\mathbb{R}^{C \times D_{m}^{g}}$ of event categories corresponding to the target dataset, where $m\in\{A,V\}$. Then, we input them into the frozen pseudo-label generator, which outputs the uni-modal event probabilities $P_{m,t}=\{P_{A,t}, P_{V,t}\}\in\mathbb{R}^{T \times C}$. Following \cite{lai2025uwav}, we also adopt soft pseudo-labels to evaluate the confidence scores, thereby reducing the interference of noise in them. The pseudo-labels are denoted as: 
\begin{equation}
\hat{P}_{m}=\text{Sigmoid}(P_{m,t}-\theta_{m})\odot Y
\end{equation}
where $\hat{P}_{m}\in\mathbb{R}^{T \times C}$ is the final AVVP pseudo-label, $\theta_{m}$ is the probability threshold, $Y$ is the video-level modality-agnostic labels used to filter out erroneous labels that include events not actually occurring in the video.

\subsection{Audio-Visual Video Parsing with Soft Constraints}
\label{softcon}

The most critical task in AVVP is to localize uni-modal events, while audio-visual events can be derived through logical operations on visual and audio events. However, existing methods mainly focus on cross-modal fusion while neglecting modality isolation, leading to ambiguous attention to uni-modal events. Therefore, we propose a soft-constrained manner that preserves uni-modal semantics while enabling effective audio-visual fusion. 

\noindent \textbf{Asymmetric Audio/Visual-Driven Fusion:}
Similar to the pseudo-label generation model in Section \ref{pretrain}, we leverage the semantic differences between temporal-aware dynamic features and temporal-unaware static features during audio-visual fusion to preserve event information in the dynamic features. Such asymmetric structure provides soft constraints on the audio and visual features, enabling the model to enhance its understanding of uni-modal events. Firstly, we perform uni-modal temporal modeling on the extracted audio and visual features $F_{m}\in\mathbb{R}^{T \times D_{m}}$($m\in\{A,V\}$), allowing the model to preliminarily focus on event-related information:
\begin{equation}
F_{m,t}=\text{FFN}[\text{MSA}(F_{m}^{q},F_{m}^{k},F_{m}^{v})]
\end{equation}
where $F_{m,t}\in\mathbb{R}^{T \times D}$ denotes temporal-aware dynamic features, $F_{m}$ represents temporal-unaware static features. Secondly, we employ an asymmetric audio/visual-driven attention for audio-visual fusion. The dynamic audio features $F_{A,t}$, which contain preliminary event semantics, can adaptively focus on information related to audio events within the static visual features $F_{V}$. Compared with direct fusion of the dynamic features $\{F_{A,t},F_{V,t}\}$, our method adopts a soft-constrained approach to avoid interference from visual semantic information on audio events, and vice versa. Each asymmetric fusion unit contains a multi-head cross-attention (MCA) and a FFN, which can be denoted as:
\begin{equation}
F_{A,t}'=\text{FFN}[\text{MCA}(F_{A,t}^{q},F_{V}^{k},F_{V}^{v})]
\end{equation}
\begin{equation}
F_{V,t}'=\text{FFN}[\text{MCA}(F_{V,t}^{q},F_{A}^{k},F_{A}^{v})]
\end{equation}
where $\{F_{A,t}',F_{V,t}'\}\in\mathbb{R}^{T \times D}$ are audio/visual-driven cross-modal features enriched with uni-modal semantic information. 

\noindent \textbf{Multi-Event Relationship Modeling:}
In addition to the dependencies between segments, dependencies also exist between events (e.g., lightning is often accompanied by rain). We propose an implicit relationship modeling module to enhance the ability of our model to infer concurrence among audio, visual, and audio-visual events. Specifically, to reduce the modality gap between the two branches and enhance the localization of events, we first perform cross-modal aggregation on $F_{m,t}'$, as follows: 
\begin{equation}
\label{Fav}
F_{AV}=\text{MLP}[\text{MCA}(F_{A,t}'^{q},F_{V,t}'^{k},F_{V,t}'^{v})]
\end{equation}
\begin{equation}
\label{Fva}
F_{VA}=\text{MLP}[\text{MCA}(F_{V,t}'^{q},F_{A,t}'^{k},F_{A,t}'^{v})]
\end{equation}
where $\{F_{AV},F_{VA}\}\in\mathbb{R}^{T \times C}$ represents the decoded audio/visual event features. Then, we employ a multi-event relationship modeling with $M$ layers to capture dependencies between events of different types:
\begin{equation}
\label{dependency}
F_{Ae}^{i}, F_{Ve}^{i}=\text{R}_{AV}^{i}[(\text{R}_{A}^{i}(F_{Ae}^{i-1}), \text{R}_{V}^{i}(F_{Ae}^{i-1})]
\end{equation}
where $\{F_{Ae}^{i}, F_{Ve}^{i}\}\in\mathbb{R}^{T \times C}$ is the output of $i$-th layer, $\text{R}_{m}^{i}$ and $\text{R}_{V}^{i}$ and $\text{R}_{AV}^{i}$ represent the relationship modeling between uni-modal events and multi-modal events, respectively. Both modules consist of a convolutional layer, an adjacency matrix $\mathcal{A}^{i}$ ($\mathcal{A}_{m}^{i}\in\mathbb{R}^{C \times C}$ for $\text{R}_{m}^{i}$ and $\mathcal{A}_{AV}^{i}\in\mathbb{R}^{2C \times 2C}$ for $\text{R}_{AV}^{i}$), batch normalization, and a LeakyReLU activation. These modules also serves as a soft constraint on uni-modal information, allowing the model to preserve uni-modal information at the event level and mitigating the ambiguous attention to uni-modal events caused by multi-modal fusion.

\noindent \textbf{Loss Functions:}
To achieve fine-grained AVVP supervision, the relation-aware features $F_{Ae}$ and $F_{Ve}$ are fed into a classifier to obtain segment-level event probabilities $\{P_{A}, P_{V}\}\in\mathbb{R}^{T \times C}$. Then, the attentive multi-modal multiple instance learning (MMIL) pooling \cite{tian2020unified} is applied to aggregate information across the temporal and modal dimensions, yielding video-level event probabilities $P$. Following \cite{lai2025uwav}, we also use the uncertainty-aware mixup loss, uncertainty-weighted classification loss, and video event classification loss to supervise our AVVP model:
\begin{equation}
\mathcal{L} = \mathcal{L}_{mix}^{A} + \mathcal{L}_{mix}^{V} + \mathcal{L}_{soft}^{A} + \mathcal{L}_{soft}^{V} + \mathcal{L}_{video}
\end{equation}
where $\mathcal{L}_{mix}^{A}$ and $\mathcal{L}_{mix}^{V}$ are self-supervised regularization terms for improving the generalization of our model, $\mathcal{L}_{soft}^{A}$ and $\mathcal{L}_{soft}^{V}$ are segment-level supervision with class-imbalance weighting, where the supervision signals are derived from the pseudo-labels generated in Section \ref{pretrain}, and $\mathcal{L}_{video}$ is the video-level supervision using coarse-grained ground truth. The full expression of these loss functions are detailed in the supplementary material. 

\section{Experiments}
\subsection{Experimental Settings}
\noindent \textbf{Dataset:}
For the pre-training of the pseudo-label generator in Stection \ref{pretrain}, we employ the dense audio-visual event dataset UnAV-100 \cite{geng2023dense}. This dataset consists of 10,790 videos with a total duration exceeding 126 hours, covering 100 categories and 30,059 audio-visual events.

We use the \emph{Look, Listen, and Parse} (LLP) \cite{tian2020unified} to evaluate the AVVP performance of different methods. This dataset serves as the sole benchmark for AVVP, containing 11,849 video clips sourced from YouTube, with a total duration of 32.9 hours. It is divided into training, validation, and test sets (10,000 videos for training, 649 for validation, and 1,200 for testing). Each video clip is 10 seconds long, covering one or more categories, with a total of 25 categories. The training set only include video-level, modality-agnostic labels, whereas the validation and testing sets include segment-level, modality-specific annotations for measuring model performance.

\noindent \textbf{Metrics:}
Following \cite{tian2020unified}, we use F-scores to evaluate audio events (A), visual events (V), and audio-visual events (AV), at both segment-level and event-level. Segment-level evaluation compares outputs and ground truth per segment, while event-level evaluation compares them per event after merging relevant segments. Additionally, we employ two metrics to assess the overall performance of the model: "Type" represents the average F-scores of A, V, and AV events, and "Event" represents the average F-scores of all events for each sample regardless of their modality. 

\noindent \textbf{Implementation Details:}
Following previous studies \cite{lai2025uwav}, the pre-trained CLIP \cite{radford2021learning} and CLAP \cite{wu2023large} are used to extract audio, visual, and textual features from the UnAV-100 dataset \cite{geng2023dense} for generator pre-training ($D_{A}'=512$, $D_{V}'=768$). For the LLP dataset \cite{tian2020unified}, we employ two sets of pre-trained backbones to extract features for experiments: (1) ResNet-152 \cite{he2016deep} + 3D ResNet \cite{tran2018closer} for visual features and VGGish \cite{hershey2017cnn} for audio features ($D_{A}=128$, $D_{V}=2560$); (2) CLIP \cite{radford2021learning} for visual features and CLAP \cite{wu2023large} for audio features ($D_{A}=D_{A}^{g}=512$, $D_{V}=D_{V}^{g}=768$). The similarity thresholds are set to $\mu_{A}=0.98$ for audio features and $\mu_{V}=0.95$ for visual features. The loss weights in Eq. (\ref{preloss}) are set to $\lambda_{A}=0.05$, $\lambda_{V}=0.15$. The number of layers in the multi-event relationship modeling in Section \ref{softcon} is set to $M=3$. During both the training of the pseudo-label generator and the AVVP model, the AdamW optimizer is used with a batch size of 64. Training lasts for 80 epochs, including a warm up stage of 10 epochs, after which a cosine annealing schedule is applied. The peak learning rate is 1e-4, and the minimum learning rates are set to 1e-5 and 5e-6, respectively. Our experiments are conducted on a single RTX 3090 GPU with 24GB of memory.

\subsection{Comparison with existing methods}
\noindent \textbf{Comparison of AVVP Performance:}
As shown in Table \ref{comp1} and Table \ref{comp2}, we compare the performance of our method with almost all existing AVVP methods using features extracted by two sets of backbone networks. Due to space limitations, we only present a subset of representative methods; the results of \cite{lin2019dual, xia2022cross, mo2022multi, yu2022mm, lin2021exploring, jiang2022dhhn, rachavarapu2023boosting, zhang2023multi, zhou2024label} are provided in our supplementary material. When using VGGish+ResNet \cite{hershey2017cnn, he2016deep, tran2018closer} for feature extraction, our method achieves either the best or the second-best performance in both segment-level and event-level video parsing. When using CLIP+CLAP \cite{radford2021learning, wu2023large} as the feature extractors, the proposed method also demonstrates excellent performance. Regardless of which set of feature extractors mentioned above is used, our method achieves new state-of-the-art average video parsing performance, reaching 63.7\% and 67.4\%, respectively. Compared with the classical method VALOR \cite{lai2023modality}, our method surpasses it by 3.2\% and 4.2\% in average performance in Table \ref{comp1} and Table \ref{comp2}, respectively. In contrast to the current state-of-the-art method UWAV \cite{lai2025uwav}, our method outperforms it by 0.9\% and 1.2\% in average performance. These results demonstrate the effectiveness and generalization of our proposed framework.

It is noteworthy that the experimental data presents two interesting phenomena: (1) Compared to UWAV \cite{lai2025uwav}, EAR achieves significant improvements in F-scores on the audio modality, while experiencing a slight decline on the visual modality. We attribute this to the visual noise introduced by EAR during the pre-training in Stage 1. Similar audio segments typically contain the same events, while due to the redundancy in visual representation, similar visual segments may exhibit significant semantic differences (for example, a scene with a dog may not include the event "barking"). Therefore, although similarity-based visual label migration can provide uni-modal supervision during pre-training, the accompanying noise can also mislead the generator. (2) Compared to TeMTG \cite{chen2025temtg} in Table \ref{comp2}, our EAR exhibits weaker performance on segment-level metrics but achieves better results on event-level metrics, especially in F-scores on audio-visual events. Because TeMTG \cite{chen2025temtg} introduces additional textual features and focuses on uni-modal multi-scale temporal modeling, it provides clearer semantic information for each segments. However, in practical inference scenarios, textual information is often unavailable. In contrast, EAR only uses visual and audio features and achieves coordinated attention to both uni-modal and multi-modal events, demonstrating strong generalization and robustness.

\begin{table*}[htbp]
  \centering
  \caption{\textbf{Comparison with state-of-the-art methods. }The visual features are extracted by ResNet-152 \cite{he2016deep} + 3D ResNet \cite{tran2018closer}, and the audio features are extracted by VGGish \cite{hershey2017cnn}. The best results are bolded, while the second-best results are underlined.}
    \begin{tabular}{c|ccccc|ccccc|c}
    \toprule
    \multirow{2.5}{*}{Method} & \multicolumn{5}{c|}{Segment-Level}    & \multicolumn{5}{c|}{Event-Level}      & \multirow{2.5}{*}{Avg.} \\
\cmidrule{2-11}          & A     & V     & AV    & Type  & Event & A     & V     & AV    & Type  & Event &  \\
    \midrule
    HAN \cite{tian2020unified}   & 60.1  & 52.9  & 48.9  & 54.0  & 55.4  & 51.3  & 48.9  & 43.0  & 47.7  & 48.0  & 51.0  \\
    MA \cite{wu2021exploring}    & 60.3  & 60.0  & 55.1  & 58.9  & 57.9  & 53.6  & 56.4  & 49.0  & 53.0  & 50.6  & 55.5  \\
    JoMoLD \cite{cheng2022joint} & 61.6  & 63.4  & 57.0  & 60.5  & 60.0  & 53.5  & 59.8  & 50.0  & 54.4  & 52.2  & 57.2  \\
    CoLeaf \cite{sardari2024coleaf} & 63.5  & 64.5  & 58.6  & 62.1  & 61.8  & 56.2  & 62.0  & 52.1  & 56.7  & 54.5  & 59.2  \\
    VAPLAN \cite{zhou2024advancing} & 62.4  & 66.7  & 60.3  & 63.1  & 61.4  & 55.7  & 63.3  & 53.7  & 57.6  & 54.3  & 59.9  \\
    LSLD \cite{fan2023revisit}  & 62.7  & 67.1  & 59.4  & 63.1  & 62.2  & 55.7  & 64.3  & 52.6  & 57.6  & 55.2  & 60.0  \\
    SDDP \cite{xie2025segment}  & 64.0  & 66.8  & 59.6  & 63.5  & 63.2  & 57.2  & 62.5  & 52.6  & 57.4  & 55.2  & 60.2  \\
    VALOR \cite{lai2023modality} & 62.8  & 66.7  & 60.0  & 63.2  & 62.3  & 57.1  & 63.9  & 54.4  & 58.5  & 55.9  & 60.5  \\
    MM-CSE \cite{zhao2025multimodal} & \underline{65.0}  & 66.8  & 60.0  & 63.9  & \underline{64.2}  & \underline{59.1}  & 64.1  & 54.9  & 59.4  & 57.6  & 61.5  \\
    PPL \cite{rachavarapu2024weakly}   & \textbf{65.9} & 66.7  & 61.9  & 64.8  & 63.7  & 57.3  & 64.3  & 54.3  & 59.9  & \underline{57.9}  & 61.7  \\
    UWAV \cite{lai2025uwav}  & 64.2  & \textbf{70.0} & \underline{63.4}  & \underline{65.9}  & 63.9  & 58.6  & \textbf{66.7} & \underline{57.5}  & \underline{60.9}  & 57.4  & \underline{62.8}  \\
    \textbf{EAR (Ours)} & \textbf{65.9}  & \underline{69.8}  & \textbf{63.8} & \textbf{66.5} & \textbf{64.8} & \textbf{60.2} & \underline{66.5}  & \textbf{58.8} & \textbf{61.8} & \textbf{58.5} & \textbf{63.7} \\
    \bottomrule
    \end{tabular}
  \label{comp1}
\end{table*}

\begin{table*}[htbp]
  \centering
  \caption{\textbf{Comparison with state-of-the-art methods. }The visual features are extracted by CLIP \cite{radford2021learning}, and the audio features are extracted by CLAP \cite{wu2023large}. The best results are bolded, while the second-best results are underlined.}
    \begin{tabular}{c|ccccc|ccccc|c}
    \toprule
    \multirow{2.5}{*}{Method} & \multicolumn{5}{c|}{Segment-Level}    & \multicolumn{5}{c|}{Event-Level}      & \multirow{2.5}{*}{Avg.} \\
\cmidrule{2-11}          & A     & V     & AV    & Type  & Event & A     & V     & AV    & Type  & Event &  \\
    \midrule
    VALOR \cite{lai2023modality} & 68.1  & 68.4  & 61.9  & 66.2  & 66.8  & 61.2  & 64.7  & 55.5  & 60.4  & 59.0  & 63.2  \\
    LINK \cite{wang2025link}  & 69.7  & 69.0  & 62.1  & 66.9  & 68.5  & 63.4  & 64.9  & 55.7  & 61.3  & 60.8  & 64.2  \\
    VAPLAN \cite{zhou2024advancing} & 69.0  & 70.2  & 63.5  & 67.6  & 67.9  & 61.9  & 66.4  & 56.9  & 61.7  & 60.1  & 64.5  \\
    LSLD \cite{fan2023revisit}  & 68.7  & 71.3  & 63.4  & 67.8  & 68.2  & 61.5  & 67.4  & 55.9  & 61.6  & 60.6  & 64.6  \\
    TIPNet \cite{zhao2025text} & 69.2  & 70.3  & 64.2  & 67.9  & 68.3  & 63.0  & 66.9  & 58.1  & 62.7  & 61.2  & 65.2  \\
    NREP \cite{jiang2024resisting}  & 70.2  & 70.9  & 64.4  & 68.5  & 68.8  & 62.8  & 67.3  & 57.6  & 62.6  & 61.1  & 65.4  \\
    MM-CSE \cite{zhao2025multimodal} & 69.5  & 71.3  & 64.2  & 68.3  & 68.9  & 63.0  & 67.5  & 57.8  & 62.7  & 61.1  & 65.4  \\
    TeMTG \cite{chen2025temtg} & \textbf{74.4} & \textbf{72.9} & 62.0  & \textbf{69.8} & \textbf{74.1} & 61.9  & \textbf{69.0} & 53.2  & 61.4  & 62.2  & 66.1  \\
    UWAV \cite{lai2025uwav}  & 68.9  & \underline{72.3}  & \underline{65.6}  & \underline{68.9}  & 68.3  & \underline{63.5}  & \underline{68.7}  & \underline{59.6}  & \underline{63.9}  & \underline{62.4}  & \underline{66.2}  \\
    \textbf{EAR (Ours)} & \underline{71.3}  & 71.8  & \textbf{66.2} & \textbf{69.8} & \underline{70.1}  & \textbf{66.4} & 68.1  & \textbf{61.4} & \textbf{65.3} & \textbf{63.9} & \textbf{67.4} \\
    \bottomrule
    \end{tabular}
  \label{comp2}
\end{table*}

\begin{table*}[htbp]
  \centering
  \caption{\textbf{Accuracy of the generated AVVP pseudo-labels on the test set. }The best results are bolded.}
    \begin{tabular}{c|ccccc|ccccc|c}
    \toprule
    \multirow{2.5}{*}{Method} & \multicolumn{5}{c|}{Segment-Level}    & \multicolumn{5}{c|}{Event-Level}      & \multirow{2.5}{*}{Avg.} \\
\cmidrule{2-11}          & A     & V     & AV    & Type  & Event & A     & V     & AV    & Type  & Event &  \\
    \midrule
    HAN+PPL \cite{rachavarapu2024weakly} & 62.5  & 55.3  & 52.3  & 56.0  & 58.3  & 55.4  & 51.1  & 46.9  & 50.9  & 50.6  & 53.9  \\
    MA+PPL \cite{rachavarapu2024weakly} & 61.7  & 61.8  & 57.5  & 60.6  & 59.4  & 55.4  & 57.9  & 51.6  & 55.0  & 52.6  & 57.4  \\
    VAPLAN \cite{zhou2024advancing} & 80.4  & 73.0  & -     & -     & -     & 71.7  & 68.3  & -     & -     & -     & - \\
    VALOR \cite{lai2023modality} & 80.4  & 71.8  & 63.7  & 72.0  & 79.7  & 72.2  & 65.9  & 55.6  & 64.6  & 68.0  & 69.4  \\
    UWAV \cite{lai2025uwav}  & 78.4  & 74.5  & 65.5  & 72.8  & 78.4  & 71.1  & 69.6  & 57.7  & 66.1  & 69.0  & 70.3  \\
    \textbf{EAR(Ours)} & \textbf{81.7} & \textbf{75.5} & \textbf{68.0} & \textbf{75.1} & \textbf{80.8} & \textbf{75.1} & \textbf{71.3} & \textbf{62.0} & \textbf{69.5} & \textbf{72.9} & \textbf{73.2} \\
    \bottomrule
    \end{tabular}
  \label{pseudo}
\end{table*}

\noindent \textbf{Accuracy of Generated Pseudo-Labels:}
For the pseudo-labels produced by our generator for the target dataset, we evaluate their accuracy on the validation and test sets. The results on the test set are shown in Table \ref{pseudo}, noting that we only present the results of open-source or well-documented methods. Due to space limitations, complete data are provided in our supplementary material. As we can see that EAR achieves the best performance across all metrics. The average video parsing performance of our generated pseudo-labels reaches a new benchmark, surpassing the current state-of-the-art methods VALOR \cite{lai2023modality} and UWAV \cite{lai2025uwav} by 3.8\% and 2.9\% on the test set, respectively. Notably, our audio pseudo-labels exhibit higher quality than the generated visual pseudo-labels. Because the features among different audio event segments are more discriminative, our similarity-based label migration can provide more accurate semantic supervision to the generator during pre-training. In summary, the proposed pseudo-label generator can achieve enhanced perception of uni-modal and multi-modal events, thus generating more accurate AVVP pseudo-labels. We will release these pseudo-labels to provide fine-grained guidance for future research. 

\begin{table}[t]
  \centering
  \caption{\textbf{Ablation study on the pseudo-label generator. }"LM" denotes our similarity-based uni-modal label migration. The best results are bolded.}
    \begin{tabular}{c|ccccc|c}
    \toprule
    Method & A     & V     & AV    & Type  & Event & Avg. \\
    \midrule
    UWAV  & 74.8  & 72.0  & 61.6  & 69.5  & 73.7  & 70.3  \\
    UWAV+LM & 76.3  & 71.4  & 61.5  & 69.8  & 75.1  & 70.8  \\
    EAR w/o LM & 77.9  & 72.0  & 63.5  & 71.1  & 76.0  & 72.1  \\
    \textbf{EAR (Ours)} & \textbf{78.4} & \textbf{73.4} & \textbf{65.0} & \textbf{72.3} & \textbf{76.8} & \textbf{73.2} \\
    \bottomrule
    \end{tabular}
  \label{pseudo_structure}
\end{table}

\begin{table}[t]
  \centering
  \caption{\textbf{Ablation study on the loss functions during pre-training. }$\{\mathcal{L}_{A}, \mathcal{L}_{V}\}$ and $\{\mathcal{L}_{AV_{1}}, \mathcal{L}_{AV_{2}}\}$ denote uni-modal are multi-modal supervision. The best results are bolded.}
    \resizebox{\columnwidth}{!}{
    \begin{tabular}{cccc|ccccc|c}
    \toprule
    $\mathcal{L}_{A}$ & $\mathcal{L}_{V}$ & $\mathcal{L}_{AV_{1}}$ & $\mathcal{L}_{AV_{2}}$ & A     & V     & AV    & Type  & Event & Avg. \\
    \midrule
    $\times$     & $\checkmark$     & $\checkmark$     & $\checkmark$     & 78.0  & 72.7  & 64.8  & 71.8  & 76.3  & 72.7  \\
    $\checkmark$     & $\times$     & $\checkmark$     & $\checkmark$     & 77.9  & 72.0  & 63.5  & 71.1  & 76.0  & 72.1  \\
    $\checkmark$     & $\checkmark$     & $\times$     & $\checkmark$     & 76.6  & 72.0  & 62.7  & 70.4  & 75.2  & 71.4  \\
    $\checkmark$     & $\checkmark$     & $\checkmark$     & $\times$     & 74.8  & 70.2  & 59.9  & 68.3  & 73.6  & 69.3  \\
    $\times$     & $\times$     & $\checkmark$     & $\checkmark$     & 76.8  & 72.1  & 63.1  & 70.7  & 75.2  & 71.6  \\
    $\checkmark$     & $\checkmark$     & $\checkmark$     & $\checkmark$     & \textbf{78.4} & \textbf{73.4} & \textbf{65.0} & \textbf{72.3} & \textbf{76.8} & \textbf{73.2} \\
    \bottomrule
    \end{tabular}}
  \label{pseudo_loss}
\end{table}

\begin{table}[t!]
  \centering
  \caption{\textbf{Ablation studies on pseudo-label performance and model performance. }The best results are bolded.}
    \begin{tabular}{c|ccccc|c}
    \toprule
    Method & A     & V     & AV    & Type  & Event & Avg. \\
    \midrule
    HAN (VALOR) & 60.0  & 65.3  & 57.2  & 60.9  & 59.1  & 60.5  \\
    HAN (UWAV) & \underline{61.4}  & \textbf{68.4} & 60.4  & 63.4  & 60.7  & 62.8  \\
    HAN (EAR) & \textbf{63.0} & 67.0  & \underline{60.6}  & \underline{63.6}  & \underline{61.4}  & \underline{63.1}  \\
    \textbf{EAR (Ours)} & \textbf{63.0} & \underline{68.2}  & \textbf{61.3} & \textbf{64.2} & \textbf{61.6} & \textbf{63.7} \\
    \bottomrule
    \end{tabular}
  \label{pseudo_han}
\end{table}

\subsection{Ablation Studies}
In this subsection, we elaborate on the effectiveness of each design component through comprehensive experimental validation. This includes ablation studies on the pseudo-label generator (in Stage 1) and the AVVP model (in Stage 3). These experiments are conducted based on frozen VGGish+ResNet \cite{hershey2017cnn, he2016deep, tran2018closer}. Note that due to space limitations, we present the average results of segment-level and event-level metrics. More details are shown in the supplementary material. 

\noindent \textbf{Ablations on the Pseudo-Label Generator:}
In Table \ref{pseudo_structure}, we validate the effectiveness of our similarity-based uni-modal label migration and asymmetric temporal modeling architecture, respectively. UWAV \cite{lai2025uwav} produces the most accurate AVVP pseudo-labels among existing methods. Its pseudo-label generator, which uses simple Transformer blocks for uni-modal temporal modeling, is supervised only by audio-visual event annotations during pre-training. (1) When the generator is supervised solely by audio-visual event annotations, our asymmetric temporal modeling architecture ("EAR w/o LM") demonstrates a marked improvement in generated pseudo-label quality on target dataset over the baseline architecture ("UWAV"). (2) When the migrated labels are introduced as uni-modal semantic guidance during pre-training ("LM"), the accuracy of the AVVP pseudo-labels generated by both UWAV \cite{lai2025uwav} and our method demonstrates substantially better results. These results indicate that our asymmetric temporal modeling and uni-modal label migration can enhance the uni-modal representations from the perspectives of model architecture and semantic guidance, respectively. By strengthening the focus on uni-modal events, EAR achieves a stronger overall capability for video semantic understanding. 

\noindent \textbf{Impact of the Pre-training Loss Functions:}
We explore the impact of loss functions during the pre-training on the accuracy of the generated pseudo-labels for the target dataset. The results are presented in Table \ref{pseudo_loss}. Without the uni-modal supervision $\mathcal{L}_{A}$ or $\mathcal{L}_{V}$ in Eq. (\ref{preloss}), the performance of the generated AVVP pseudo-labels shows some degradation compared with the generator trained with the complete loss function, while the removal of both results in pseudo-labels of lower quality. Moreover, removing the multi-modal constraints $\mathcal{L}_{AV_{1}}$ or $\mathcal{L}_{AV_{2}}$ has a greater effect on the pseudo-label generator, leading to insufficient event attention, especially for audio-visual events. By contrast, applying both uni-modal and multi-modal semantic guidance to the two temporal modeling branches simultaneously can enhance the perception of audio, visual, and audio-visual events by our pseudo-label generator. 

\begin{table}[t!]
  \centering
  \caption{\textbf{Ablation study on the AVVP model with soft constraints. }"AMDF" denotes the Asymmetric Audio/Visual-Driven Fusion module, and "ERM" denotes the Multi-Event Relationship Modeling module. The best results are bolded.}
    \begin{tabular}{cc|ccccc|c}
    \toprule
    AMDF  & ERM   & A     & V     & AV    & Type  & Event & Avg. \\
    \midrule
    $\times$     & $\checkmark$     & 62.1  & 66.6  & 59.4  & 62.7  & 60.7  & 62.3  \\
    $\checkmark$     & $\times$     & 62.1  & 67.2  & 60.4  & 63.2  & 61.0  & 62.8  \\
    $\checkmark$     & $\checkmark$     & \textbf{63.0} & \textbf{68.2} & \textbf{61.3} & \textbf{64.2} & \textbf{61.6} & \textbf{63.7} \\
    \bottomrule
    \end{tabular}
  \label{avvp_structure}
\end{table}

\begin{table}[t!]
  \centering
  \caption{\textbf{Ablation study of different structure in the Asymmetric Audio/Visual-Driven Fusion module. } The best results are bolded.}
    \begin{tabular}{c|ccccc|c}
    \toprule
    Method & A     & V     & AV    & Type  & Event & Avg. \\
    \midrule
    MSA+MCA & 62.3  & 66.9  & 60.1  & 63.1  & 60.8  & 62.7  \\
    HAN   & 62.4  & 67.4  & 60.5  & 63.4  & 61.2  & 63.0  \\
    \textbf{AMDF(Ours)} & \textbf{63.0} & \textbf{68.2} & \textbf{61.3} & \textbf{64.2} & \textbf{61.6} & \textbf{63.7} \\
    \bottomrule
    \end{tabular}
  \label{avvp_amdf}
\end{table}

\begin{table}[t!]
  \centering
  \caption{\textbf{Ablation study of different structure in the Multi-Event Relationship Modeling module. } The best results are bolded, while the second-best results are underlined.}
    \begin{tabular}{c|ccccc|c}
    \toprule
    Method & A     & V     & AV    & Type  & Event & Avg. \\
    \midrule
    Attention & 61.9  & 66.1  & 59.6  & 62.5  & 60.4  & 62.1  \\
    ERM A/V-AV & \textbf{63.1} & \underline{67.6}  & \underline{60.9}  & \underline{63.9}  & \textbf{61.7} & \underline{63.4}  \\
    ERM w/o A/V & 62.3  & 67.0  & 60.3  & 63.2  & 61.0  & 62.8  \\
    ERM w/o AV & 63.0  & 66.7  & 60.7  & 63.5  & \underline{61.6}  & 63.1  \\
    \textbf{ERM (Ours)} & \underline{63.0}  & \textbf{68.2} & \textbf{61.3} & \textbf{64.2} & \underline{61.6}  & \textbf{63.7} \\
    \bottomrule
    \end{tabular}
  \label{avvp_erm}
\end{table}

\begin{figure*}[t]
	\centering
	\includegraphics[width=17cm,height=5.417cm]{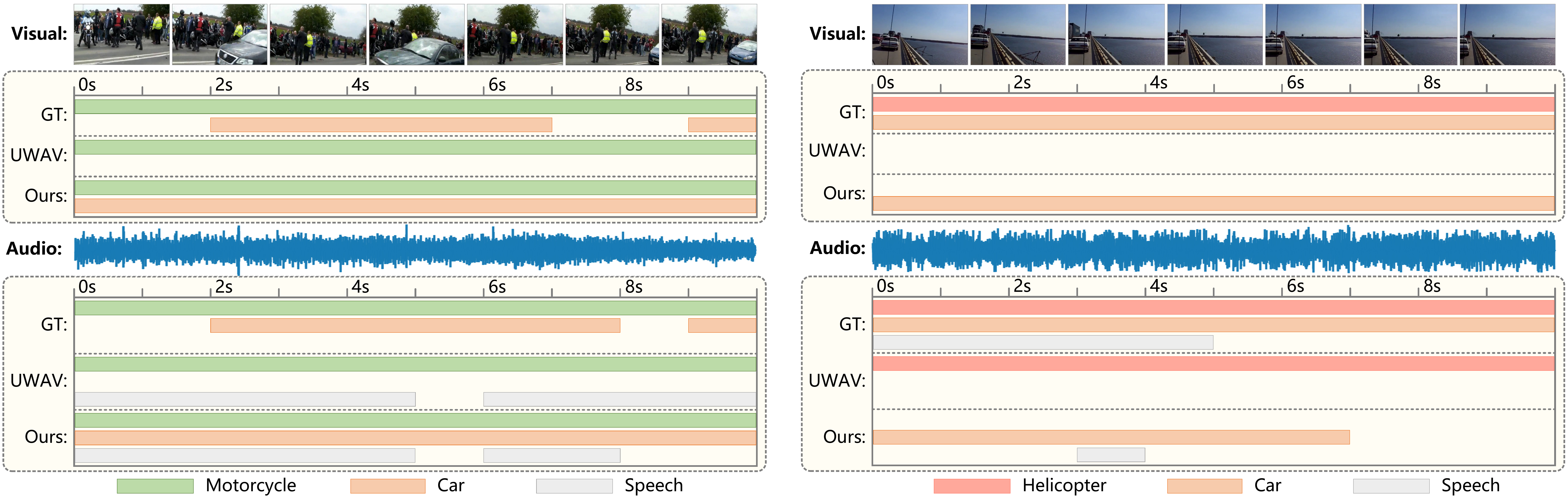}
	\caption{Qualitative comparison of audio-visual video parsing with state-of-the-art methods. "GT" denotes the ground truth.}
	\label{qualitative}
\end{figure*}

\noindent \textbf{Ablations on the pseudo-labels and model performance:}
To further validate the effectiveness of our pseudo-labels in weakly-supervised AVVP training, we conduct experiments using pseudo-labels generated by different methods on a unified baseline model. The results are presented in Table \ref{pseudo_han}. "HAN(VALOR)", "HAN(UWAV)", and "HAN(EAR)" denote training HAN \cite{tian2020unified} with pseudo-labels generated by VALOR \cite{lai2023modality}, UWAV \cite{lai2025uwav}, and our EAR, respectively. First, the AVVP models trained with pseudo-labels generated by UWAV \cite{lai2025uwav} or EAR (that consider inter-segment dynamic relationships) significantly outperform that trained with the static pseudo-labels generated by VALOR \cite{lai2023modality}. Second, HAN \cite{tian2020unified} trained with pseudo-labels from EAR outperforms the version trained with pseudo-labels from UWAV \cite{lai2025uwav}, indicating the effectiveness of our enhancement of uni-modal representations during pre-training. Finally, the superiority of our AVVP model is verified as EAR achieves higher performance than HAN \cite{tian2020unified} when trained on the same pseudo-labels. This suggests that our model with soft constraints can better capture uni-modal and multi-modal events in videos. 

\noindent \textbf{Ablations on the AVVP Model:}
To validate the role of the modules in Stection \ref{softcon} (Stage 3), we perform ablation studies on our AVVP model, and the results are shown in Table \ref{avvp_structure}. Without the asymmetric audio/visual-driven fusion module, the video parsing performance of our model decreases significantly, demonstrating the importance of the early feature-level temporal modeling. Removing the multi-event relationship modeling module also negatively impacts our AVVP model, which demonstrates the effectiveness of the event-level dependency modeling. Overall, it is the integration of low-level temporal modeling with cross-modal fusion and high-level semantic modeling that enhances the audio-visual video parsing capability of our model. 

\noindent \textbf{Impact of different structures in AMDF:}
In Table \ref{avvp_amdf}, we replace our asymmetric audio/visual-driven fusion module with other classical structures to investigate the effect of different cross-modal temporal modeling approaches on AVVP model performance. "MSA+MCA" denotes sequentially applying uni-modal self-attention and multi-modal cross-attention; "HAN" denotes employing the hybrid attention network from \cite{tian2020unified} for temporal modeling and audio-visual fusion. We observe that the model equipped with our asymmetric audio/visual-driven fusion module outperforms the model equipped with either "MSA+MCA" or "HAN" across all metrics, achieving an average video parsing performance of 63.7\%. This indirectly indicates that the asymmetric soft-constrained fusion can effectively extract cross-modal event information while preserving uni-modal semantics and avoiding interference between them. 

\noindent \textbf{Impact of different structures in ERM:}
To further explore the performance of our multi-event relationship modeling module, we apply different structures to model the dependencies among multiple events. The results are shown in Table \ref{avvp_erm}. Among them, "Attention" denotes directly concatenating the decoded event features $F_{AV}$ and $F_{VA}$ from Eq. (\ref{Fav}) and Eq. (\ref{Fva}), and then using self-attention to capture multi-modal multi-event relationships. "ERM A/V-AV" denotes that the structure in Eq. (\ref{dependency}) is modified to first perform $M$ layers of uni-modal event relationship modeling, followed by $M$ layers of multi-modal event relationship modeling. "ERM w/o A/V" and "ERM w/o AV" denote the removal of uni-modal event modeling and multi-modal event modeling from our method, respectively. It can be seen that, compared to other structures, "ERM" and "ERM A/V-AV" achieve the best or second-best performance across all metrics, with average video parsing performance reaching 63.7\% and 63.4\%, respectively. This suggests that modeling the dependencies of both uni-modal and multi-modal events has a positive impact on the performance of our model. Moreover, compared to "ERM A/V-AV", our module achieves slightly better AVVP performance. Because our module alternately conducts uni-modal and multi-modal event relationship modeling, it enables the model to preserve uni-modal event information effectively. 

\begin{figure}[t]
	\centering
	\includegraphics[width=8.8cm,height=4.32cm]{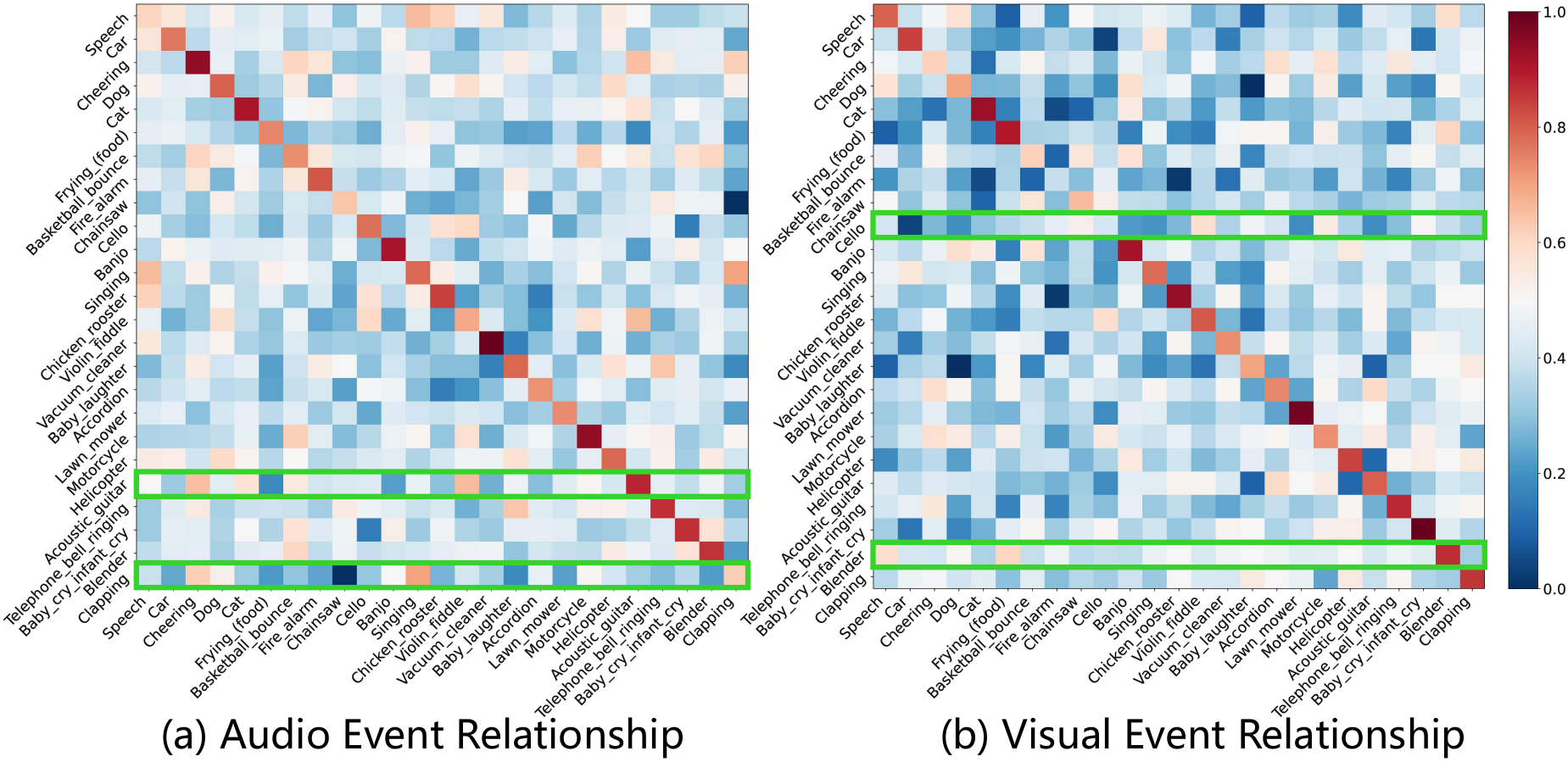}
	\caption{Visualization of uni-modal event dependencies learned by our Multi-Event Relationship Modeling.}
	\label{MER-uni}
\end{figure}

\begin{figure}[t]
	\centering
	\includegraphics[width=8.4cm,height=7.56cm]{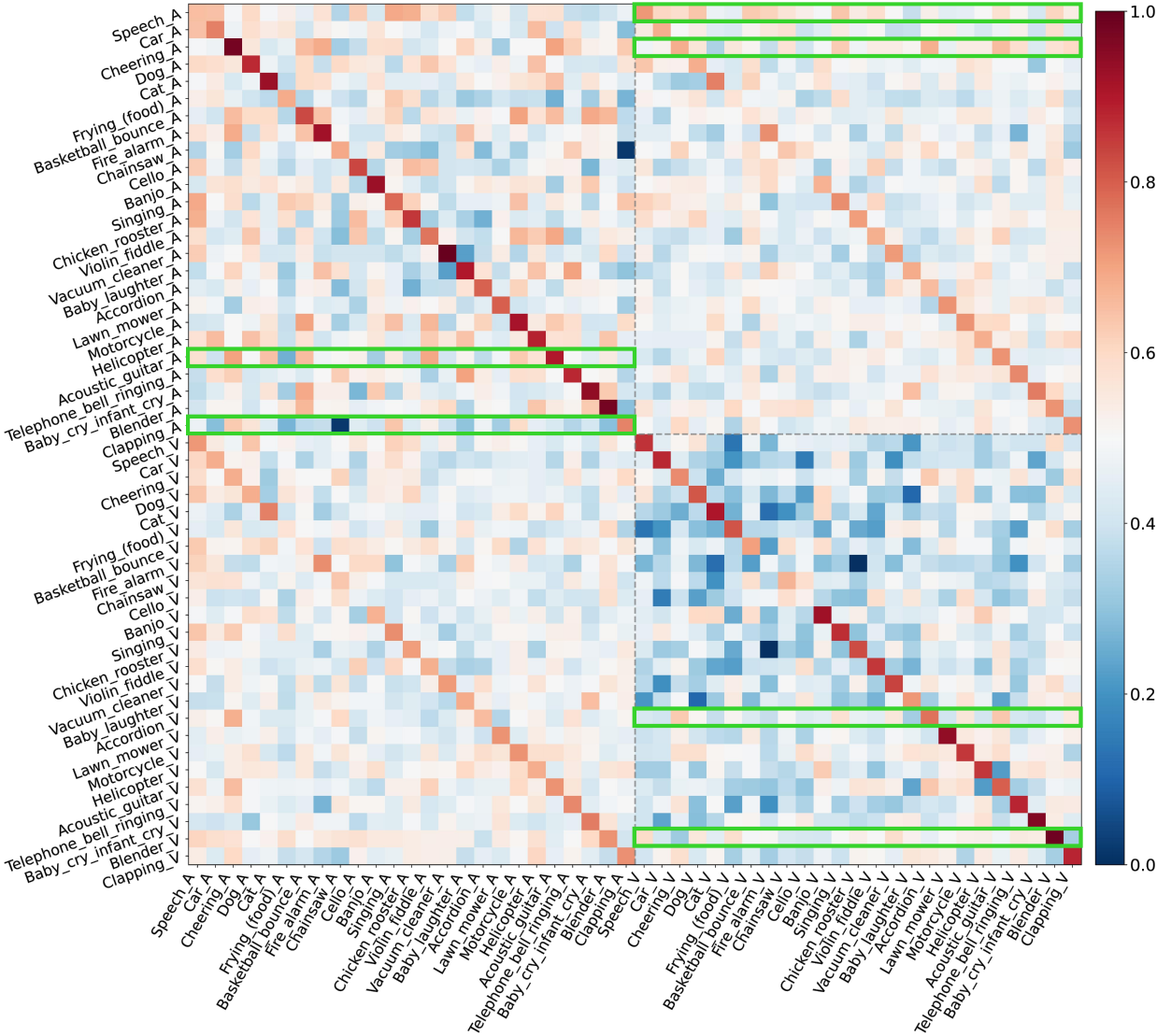}
	\caption{Visualization of multi-modal event dependencies learned by our Multi-Event Relationship Modeling.}
	\label{MER-multi}
\end{figure}

\subsection{Qualitative Results}
\noindent \textbf{Qualitative Comparison of AVVP:}
To demonstrate the performance advantage of EAR, we present the predictions of our method and the state-of-the-art method on the audio-visual video parsing task, with the visualization results shown in Figure \ref{qualitative}. In the left example, compared to UWAV \cite{lai2025uwav}, our method does not miss the uni-modal event "Car". Although both methods output the erroneous event "Speech" in the audio modality, which is not present in the ground truth, EAR has a lower error rate. In the right example, our prediction results are also more accurate, with a lower proportion of missed detections. In contrast, UWAV \cite{lai2025uwav} only parses the audio event of "helicopter" and fails to detect all other events. These results indicate that our method can better focus on uni-modal event information in the AVVP task, enabling accurate perception of uni-modal events. However, EAR also exhibits instances of over detection or under detection, which may be due to the negative impact of noise in our pseudo-labels. How to improve the semantic guidance during the pre-training of the pseudo-label generator to improve the quality of AVVP pseudo-labels remains a critical issue. 

\noindent \textbf{Visualization of learned event dependencies:}
To further verify the high-level semantic modeling capability of the multi-event relationship modeling module, we visualize the captured dependencies for uni-modal and multi-modal events, with the results shown in Figure \ref{MER-uni} and Figure \ref{MER-multi}, respectively. Representative examples in the figures are marked with green bounding boxes. \textbf{(1)} For uni-modal dependencies, we aggregate the adjacency matrices $\mathcal{A}_{m}^{i}$ from $M$ layers and average the upper and lower triangular parts to reduce bias. We can observe that the audio event "Acoustic\_guitar" exhibits strong relationships with "Cheering" and "Violin\_fiddle". Meanwhile, the audio event "Clapping" has a strong dependency with "Singing" and the weakest dependency with "Chainsaw", which aligns with common scenarios. Certain dependencies also exist among visual events. For instance, "Cello" and "Violin\_fiddle" exhibit a strong association, while "Blender" shows strong relationships with "Violin\_fiddle". Notably, the overall associations among visual events are relatively weak compared to audio associations, due to the limitation of original image size that restricts multiple entities from appearing in a single frame, whereas multiple sounds can naturally coexist in the audio track. \textbf{(2)} For multi-modal dependencies, we aggregate $\mathcal{A}_{AV}^{i}$ and $\mathcal{A}_{m}^{i}$ from $M$ layers, and average the upper and lower triangular parts to obtain a comprehensive visualization of event dependencies. The upper-right part of Figure \ref{MER-multi} shows that the audio event "Speech" has noticeable associations with many visual events, consistent with the presence of substantial background speech in the original video. The audio event "Cheering" shows clear associations with visual events "Accordion" and "Acoustic\_guitar", which also aligns with intuitive understanding. Second, the upper-left and lower-right parts of Figure \ref{MER-multi} show similar results to Figure \ref{MER-uni}, which indirectly demonstrates the effectiveness of our soft constraints in preserving uni-modal information. In summary, these results show that our multi-event relationship modeling can capture dependencies among concurrent events, while enhancing the attention on uni-modal events to prevent semantic ambiguity caused by cross-modal fusion. 

\section{Conclusion}
In this paper, we propose an enhanced uni-modal representation framework to improve the audio-visual video parsing capability of the model. First, we apply a similarity-based label migration method, introducing uni-modal supervision for the pseudo‑label generator to achieve semantic-level uni-modal enhancement. Second, we design modules with soft constraints, including the asymmetric audio/visual-driven fusion and multi-event relationship modeling, to enhance the focus of our AVVP model on uni-modal events from an architectural perspective. These designs enable EAR to coordinate its attention across uni-modal and multi-modal events, thus delivering state-of-the-art performance. The accuracy of our fine-grained pseudo-labels and the overall AVVP performance surpass all previous works, demonstrating the effectiveness of our framework. 

\noindent \textbf{Limitations:}
Although our similarity-based label migration enhances semantic guidance for the pseudo-label generator during pre-training, the migrated labels contain some noise, particularly in the visual modality. Because CLIP and CLAP encoders focus on holistic information and struggle to capture details in the original data. Employing an encoders with multi-scale analysis capabilities may mitigate this issue. 

\section*{Acknowledgments}
This work was supported by the National Key R\&D Program of China (2024YFC3015604), the Beijing Natural Science Foundation under Grant F2024203115, and the China Postdoctoral Science Foundation under Grant Number 2024M750255.


{
\bibliographystyle{IEEEtran}
\bibliography{ref}
}
%
%
%
%
%
%
%
%
%
%

\begin{IEEEbiography}[{\includegraphics[width=1in,height=1.25in,clip,keepaspectratio]{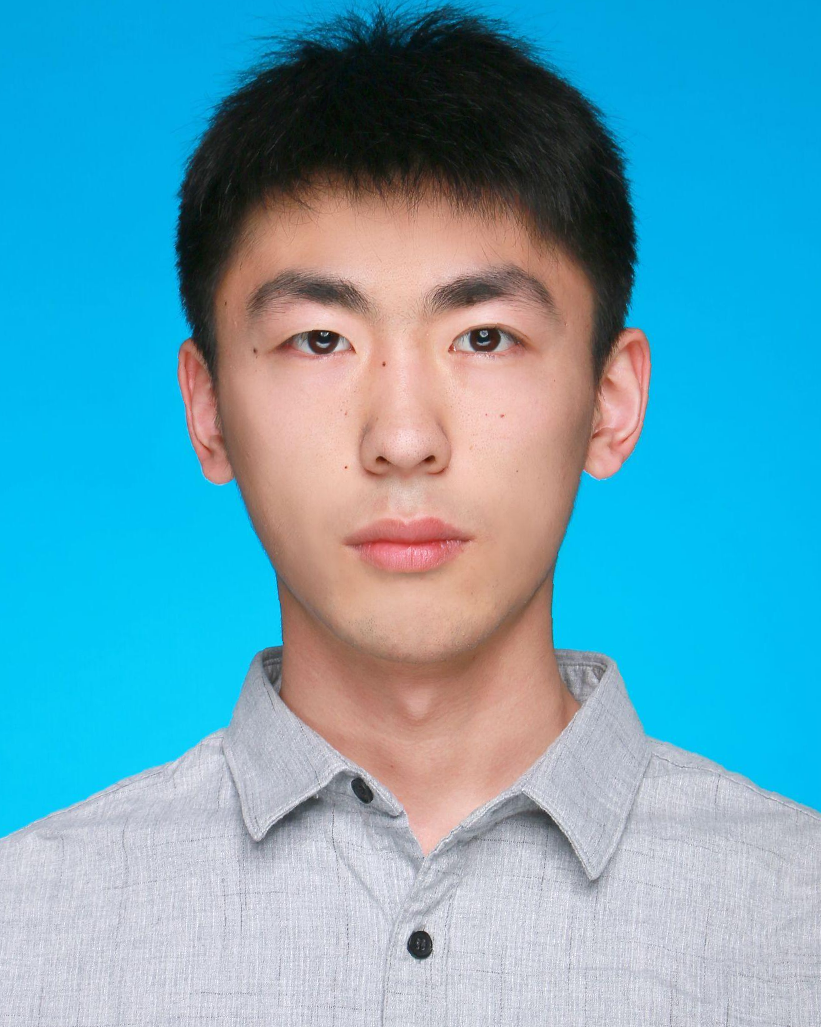}}]{Huilai Li}
received the B.Eng. degree in 2021 from the Zhengzhou University, Zhengzhou, China. He is currently working toward the Ph.D. degree with the School of Intelligent Engineering and Automation, Beijing University of Posts and Telecommunications, Beijing, China. His research interests include multi-modal learning, computer vision, and deep learning.
\end{IEEEbiography}

\begin{IEEEbiography}[{\includegraphics[width=1in,height=1.25in,clip,keepaspectratio]{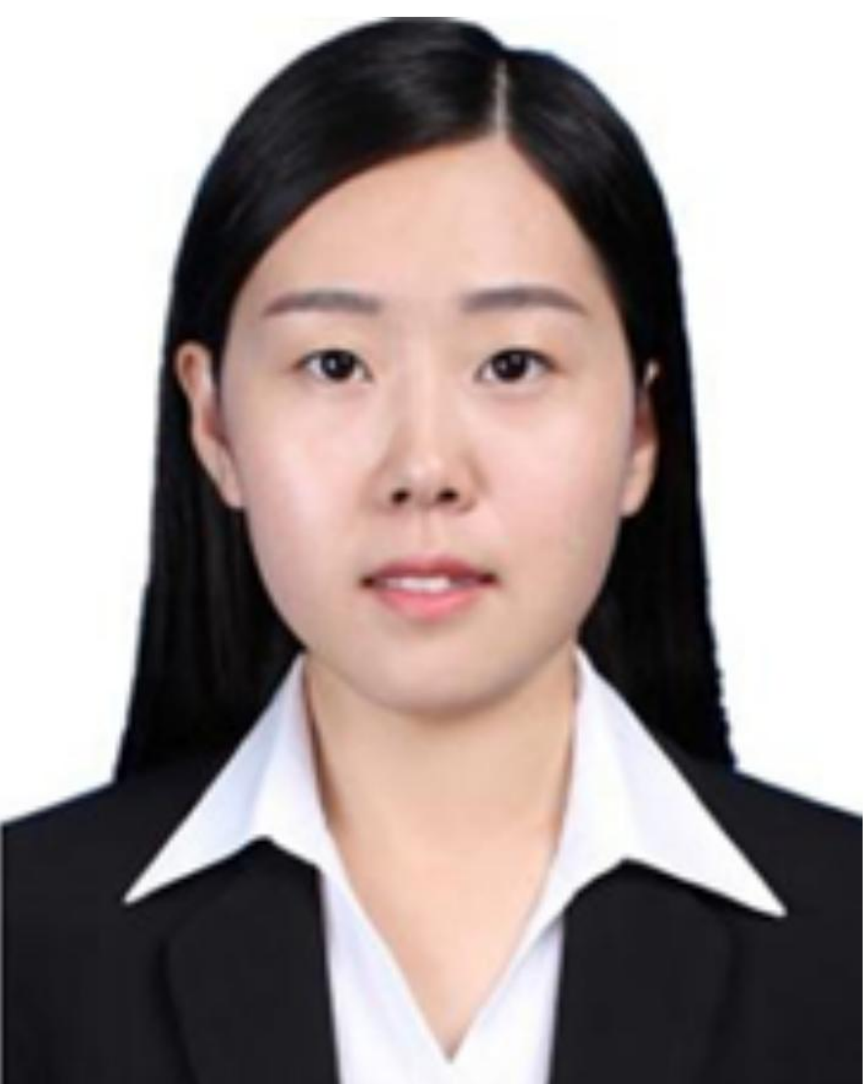}}]{Xiaomeng Di}
received the M.S. degree in Computer Science and Technology from University of Science and Technology Beijing in 2020 and joined State Grid Corporation of China in the same year. Since assuming the role of Engineer in 2023, she has focused on areas such as Artificial Intelligence and Intelligent Analysis.
\end{IEEEbiography}

\begin{IEEEbiography}[{\includegraphics[width=1in,height=1.25in,clip,keepaspectratio]{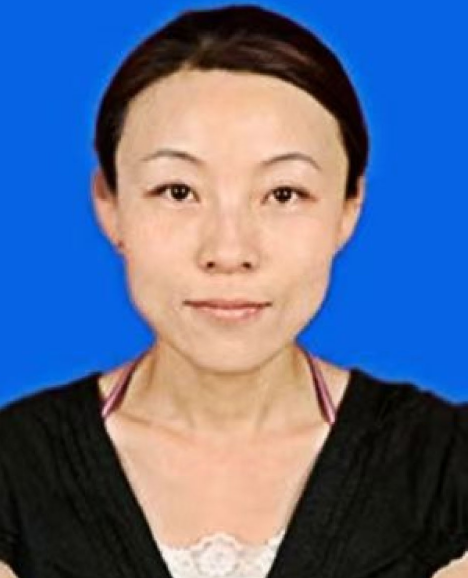}}]{Ying Xing} 
received the Ph.D. degree in Computer Science and Technology from Beijing University of Posts and Telecommunications, Beijing, China in 2014. She is currently an associate professor with the School of Intelligent Engineering and Automation, Beijing University of Posts and Telecommunications. Her research interests include software testing and vulnerability detection.
\end{IEEEbiography}

\begin{IEEEbiography}[{\includegraphics[width=1in,height=1.25in,clip,keepaspectratio]{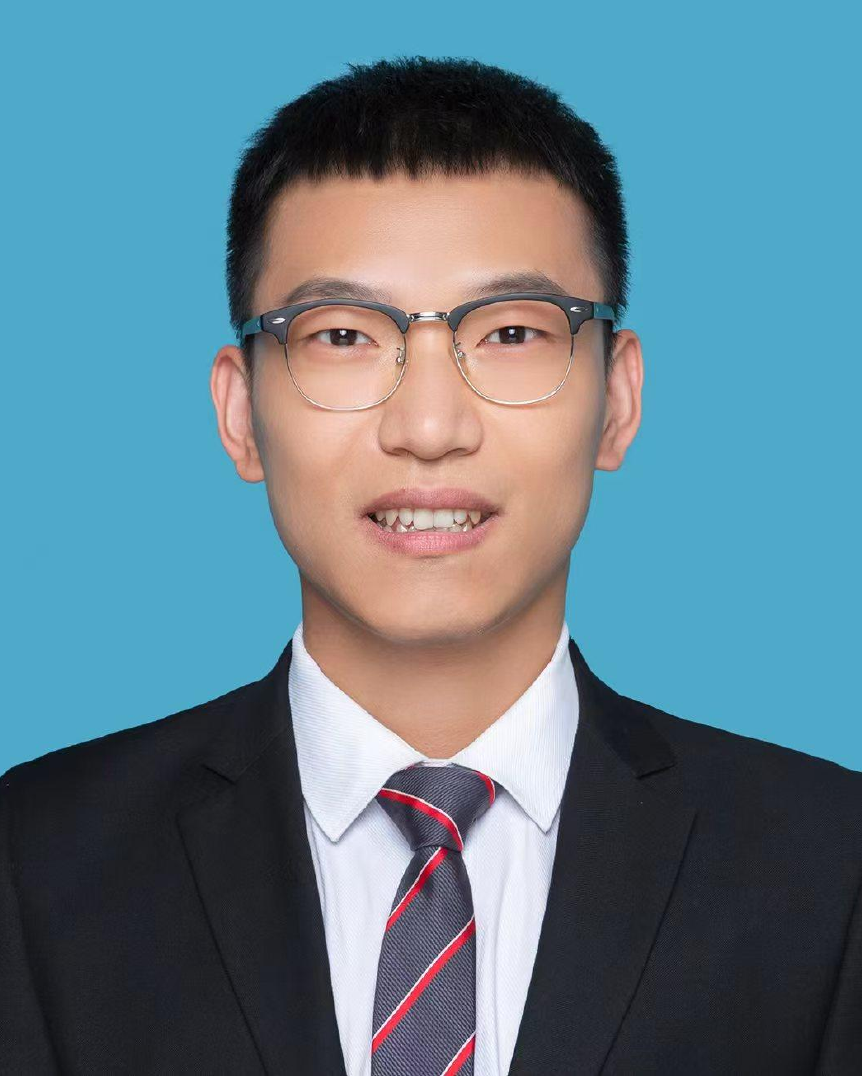}}]{Yonghao Dang}
received Bachelor degree in computer science and technology from the University of Jinan, Jinan, China, in 2018, and the Ph.D. degree from the School of Artificial Intelligence, Beijing University of Posts and Telecommunications, Beijing, China, in 2023. His research interests include computer vision, image processing, and deep learning.
\end{IEEEbiography}

\begin{IEEEbiography}[{\includegraphics[width=1in,height=1.25in,clip,keepaspectratio]{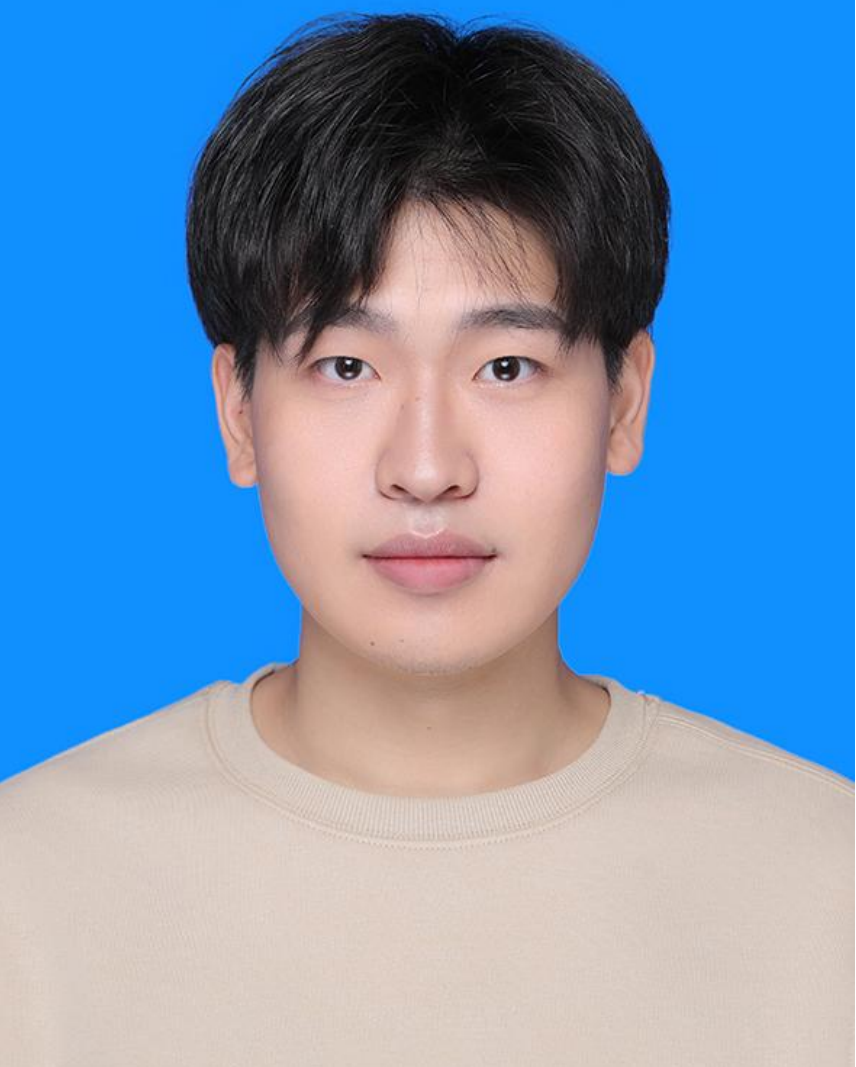}}]{Yiming Wang}
graduated from Hebei GEO University with a Bachelor degree, Shijiazhuang, China. He is currently working toward the Ph.D. degree with the School of Intelligent Engineering and Automation, Beijing University of Posts and Telecommunications, Beijing, China. His research interests include multi-modal learning, computer vision, and deep learning.
\end{IEEEbiography}

\begin{IEEEbiography}[{\includegraphics[width=1in,height=1.25in,clip,keepaspectratio]{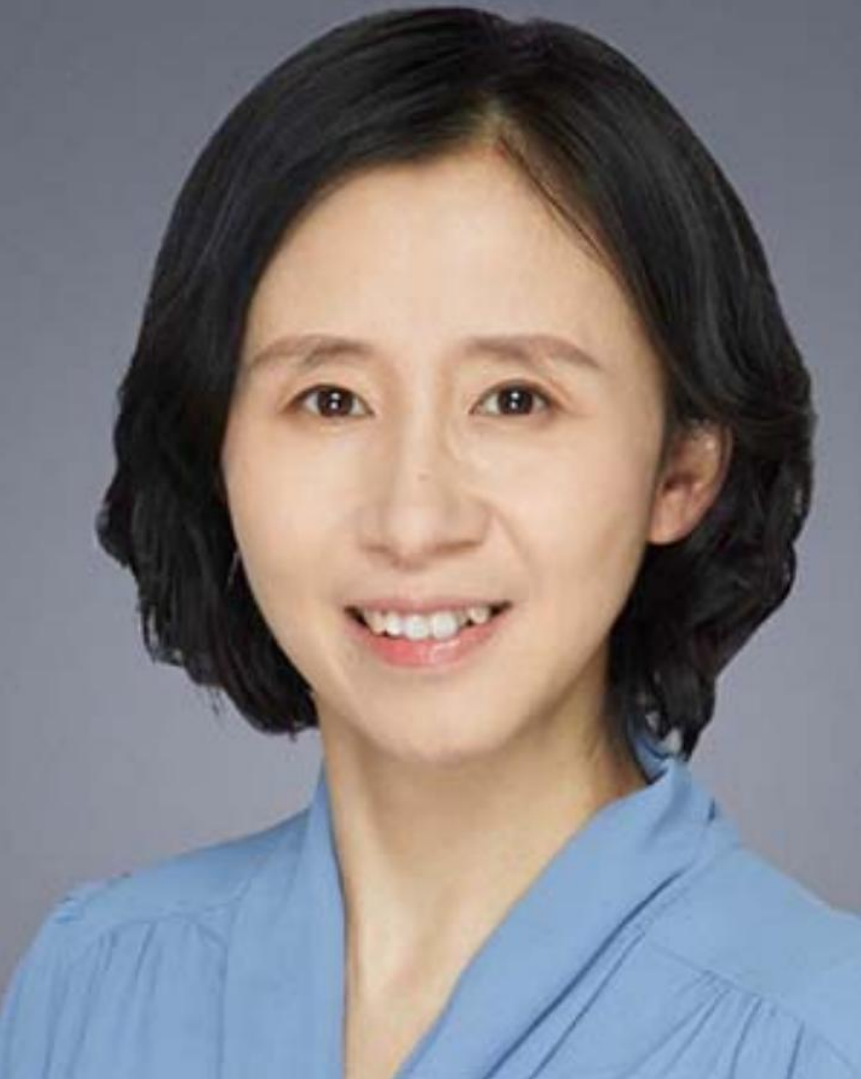}}]{Jianqin Yin}
(Member, IEEE) received the Ph.D. degree from Shandong University, Jinan, China, in 2013. She is currently a Professor with the School of Intelligent Engineering and Automation, Beijing University of Posts and Telecommunications, Beijing, China. Her research interests include service robots, pattern recognition, machine learning, and image processing.
\end{IEEEbiography}
%
%
%
%
%

\clearpage
\section*{Supplementary Material}
In the supplementary material, we provide detailed information about the proposed framework, including Methodology Details (\emph{Post-Processing of Uni-Modal Label Migration} and \emph{Loss Functions}) and Experimental Details (\emph{Comparative Studies}, \emph{Ablation Studies} and \emph{Qualitative Results}).

\renewcommand{\thesection}{S-\Roman{section}}
\setcounter{section}{5}
\section{Methodology Details}
\subsection{Post-Processing of Uni-Modal Label Migration}
Our Similarity-Based Uni-Modal Label Migration may suffer from duplicate annotations due to multiple similar segments containing the same audio-visual events. An example is provided in Figure \ref{Spost}, where the second row of the high similarity matrix is multiplied by the second column of the audio-visual label (blue box), resulting in a value greater than 1 (1.8), which does not meet the labeling requirements. Because multiple segments similar to the second segment contain the same audio-visual label, as indicated by the orange box in Figure \ref{Spost}. Therefore, we need to perform post-processing on the migrated uni-modal labels by averaging duplicate annotations, while ground truth annotations originally labeled as 1 are kept unchanged (i.e., not averaged). 
\renewcommand{\thefigure}{S-\arabic{figure}}
\setcounter{figure}{5}
\begin{figure}[h]
	\centering
	\includegraphics[width=6cm,height=2.494cm]{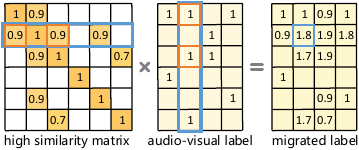}
	\caption{Example of duplicate annotation in label migration.}
	\label{Spost}
\end{figure}

Specifically, we divide the high similarity matrix $\hat{S}_{m}^{p}$ into row vectors and the audio-visual label $Y_{AV}^{p}$ into column vectors:
\renewcommand{\theequation}{S-\arabic{equation}}
\setcounter{equation}{16}
\begin{equation}
\hat{S}_{m}^{p} = [\hat{s}_{1}^{p}, \hat{s}_{2}^{p}, \dots, \hat{s}_{T'}^{p}]^{\top}, 
Y_{AV}^{p} = [y_{1}^{p}, y_{2}^{p}, \dots, y_{C'}^{p}]
\end{equation}

The migrated uni-modal label $Y_{ms}^{p}$ in Eq. (2) is denoted as: 
\begin{equation}
Y_{ms}^{p} = 
\begin{pmatrix}
\hat{s}_{1}^{p} y_{1}^{p} & \hat{s}_{1}^{p} y_{2}^{p} & \dots & \hat{s}_{1}^{p} y_{C'}^{p} \\[3pt]
\hat{s}_{2}^{p} y_{1}^{p} & \hat{s}_{2}^{p} y_{2}^{p} & \dots & \hat{s}_{2}^{p} y_{C'}^{p} \\[3pt]
\vdots & \vdots & \ddots & \vdots \\[3pt]
\hat{s}_{T'}^{p} y_{1}^{p} & \hat{s}_{T'}^{p} y_{2}^{p} & \dots & \hat{s}_{T'}^{p} y_{C'}^{p}
\end{pmatrix}
\end{equation}

Second, we calculate the number of duplicate annotations: 
\begin{equation}
D_{n} = M_{m}^{p}Y_{AV}^{p}
\end{equation}
where $D_{n}$ denotes the number of duplicate annotations for each migrated label, $M_{m}^{p}$ is the high similarity mask matrix from Eq. (1) in the main text. Then, the duplicate annotations are averaged based on $D_{n}$: 
\begin{equation}
Y_{ms}^{D'} = Y_{ms}^{p} \oslash D_{n}
\end{equation}	
where $\oslash$ denotes element-wise division. Finally, to preserve the original audio-visual ground truth $Y_{AV}^{p}$ unchanged, we we take the element-wise maximum between $Y_{ms}^{D'}$ and $Y_{AV}^{p}$:
\begin{equation}
Y_{ms}^{D} = \text{max}(Y_{ms}^{D'}, Y_{AV}^{p})
\end{equation}
where $Y_{ms}^{D}\in\mathbb{R}^{T' \times C'}$ is the final migrated label. At this point, the post-processing of the migrated uni-modal labels is completed, ensuring that sss complies with the labeling requirements. Additionally, most of the migrated labels in aaa are assigned values between 0 and 1 rather than directly set to 1, which can serve as confidence scores for the migrated labels to mitigate the negative impact of noise.

\renewcommand{\thetable}{S-\Roman{table}}
\setcounter{table}{9}
\begin{table*}[htbp]
  \centering
  \caption{\textbf{Comparison with state-of-the-art methods. }The visual features are extracted by ResNet-152 \cite{he2016deep} + 3D ResNet \cite{tran2018closer}, and the audio features are extracted by VGGish \cite{hershey2017cnn}. The best results are bolded, while the second-best results are underlined.}
    \begin{tabular}{c|ccccc|ccccc|c}
    \toprule
    \multirow{2.5}{*}{Method} & \multicolumn{5}{c|}{Segment-Level}    & \multicolumn{5}{c|}{Event-Level}      & \multirow{2.5}{*}{Avg.} \\
\cmidrule{2-11}          & A     & V     & AV    & Type  & Event & A     & V     & AV    & Type  & Event &  \\
    \midrule
    AVE \cite{tian2018audio}   & 49.9  & 37.3  & 37.0  & 41.4  & 43.6  & 43.6  & 32.4  & 32.6  & 36.2  & 37.4  & 39.1  \\
    AVSDN \cite{lin2019dual} & 47.8  & 52.0  & 37.1  & 45.7  & 50.8  & 34.1  & 46.3  & 26.5  & 35.6  & 37.7  & 41.4  \\
    HAN \cite{tian2020unified}   & 60.1  & 52.9  & 48.9  & 54.0  & 55.4  & 51.3  & 48.9  & 43.0  & 47.7  & 48.0  & 51.0  \\
    CMBS \cite{xia2022cross}  & 60.2  & 54.3  & 50.0  & 54.8  & 55.7  & 51.1  & 50.8  & 43.7  & 48.5  & 48.3  & 51.7  \\
    MGN \cite{mo2022multi}   & 60.8  & 55.4  & 50.4  & 55.5  & 57.2  & 51.1  & 52.4  & 44.4  & 49.3  & 49.1  & 52.6  \\
    CM-PIE \cite{chen2024cm} & 61.7  & 55.2  & 50.1  & 55.7  & 56.8  & 53.7  & 51.3  & 43.6  & 49.5  & 51.3  & 52.9  \\
    MA \cite{wu2021exploring}    & 60.3  & 60.0  & 55.1  & 58.9  & 57.9  & 53.6  & 56.4  & 49.0  & 53.0  & 50.6  & 55.5  \\
    MM-pyramid \cite{yu2022mm} & 61.1  & 60.3  & 55.8  & 59.7  & 59.1  & 53.8  & 56.7  & 49.4  & 54.1  & 51.2  & 56.1  \\
    CVCMS \cite{lin2021exploring} & 60.8  & 63.5  & 57.0  & 60.5  & 59.5  & 53.8  & 58.9  & 49.5  & 54.0  & 52.1  & 57.0  \\
    JoMoLD \cite{cheng2022joint} & 61.6  & 63.4  & 57.0  & 60.5  & 60.0  & 53.5  & 59.8  & 50.0  & 54.4  & 52.2  & 57.2  \\
    VAPLAN \cite{zhou2024advancing} & 62.4  & 66.7  & 60.3  & 63.1  & 61.4  & 55.7  & 63.3  & 53.7  & 57.6  & 54.3  & 59.9  \\
    LGFNet \cite{sun2024multi} & 62.2  & 64.0  & 56.9  & 61.0  & 60.5  & 54.4  & 60.3  & 49.8  & 54.8  & 52.9  & 57.7  \\
    DHHN \cite{jiang2022dhhn}  & 61.7  & 63.2  & 56.8  & 60.6  & 59.7  & 54.8  & 60.4  & 51.1  & 55.4  & 53.3  & 57.7  \\
    Rachavarapu \emph{et al.} \cite{rachavarapu2023boosting} & 63.1  & 63.5  & 57.7  & 61.4  & 60.6  & 54.1  & 60.3  & 51.5  & 55.2  & 52.3  & 58.0  \\
    Xu \emph{et al.} \cite{xu2024rethink} & 61.9  & 64.8  & 57.6  & 61.4  & 60.9  & 53.9  & 61.6  & 50.2  & 55.2  & 53.1  & 58.1  \\
    CMRN \cite{li2024exploring}  & 63.2  & 63.9  & 57.0  & 61.4  & 61.4  & 55.7  & 60.4  & 49.3  & 55.1  & 54.2  & 58.2  \\
    AVFAS \cite{zhang2023multi} & 61.4  & 64.5  & 58.4  & 61.4  & 59.9  & 54.2  & 61.9  & 51.9  & 56.0  & 53.0  & 58.3  \\
    CoLeaf \cite{sardari2024coleaf} & 63.5  & 64.5  & 58.6  & 62.1  & 61.8  & 56.2  & 62.0  & 52.1  & 56.7  & 54.5  & 59.2  \\
    CMPAE \cite{gao2023collecting} & 64.2  & 66.4  & 59.2  & 63.3  & 62.8  & 56.6  & 63.7  & 51.8  & 51.9  & 55.7  & 59.6  \\
    LEAP (HAN) \cite{zhou2024label} & 62.7  & 65.6  & 59.3  & 62.5  & 61.8  & 56.4  & 63.1  & 54.1  & 57.8  & 55.0  & 59.8  \\
    LSLD \cite{fan2023revisit}  & 62.7  & 67.1  & 59.4  & 63.1  & 62.2  & 55.7  & 64.3  & 52.6  & 57.6  & 55.2  & 60.0  \\
    SDDP \cite{xie2025segment}  & 64.0  & 66.8  & 59.6  & 63.5  & 63.2  & 57.2  & 62.5  & 52.6  & 57.4  & 55.2  & 60.2  \\
    VALOR \cite{lai2023modality} & 62.8  & 66.7  & 60.0  & 63.2  & 62.3  & 57.1  & 63.9  & 54.4  & 58.5  & 55.9  & 60.5  \\
    MM-CSE \cite{zhao2025multimodal} & \underline{65.0}  & 66.8  & 60.0  & 63.9  & \underline{64.2}  & 59.1  & 64.1  & 54.9  & 59.4  & 57.6  & 61.5  \\
    PPL \cite{rachavarapu2024weakly}   & \textbf{65.9} & 66.7  & 61.9  & 64.8  & 63.7  & 57.3  & 64.3  & 54.3  & 59.9  & \underline{57.9}  & 61.7  \\
    LEAP (MM-pyramid) \cite{zhou2024label} & 64.8  & 67.7  & 61.8  & 64.8  & 63.6  & \underline{59.2}  & 64.9  & 56.5  & 60.2  & 57.4  & 62.1  \\
    UWAV \cite{lai2025uwav}  & 64.2  & \textbf{70.0} & \underline{63.4}  & \underline{65.9}  & 63.9  & 58.6  & \textbf{66.7} & \underline{57.5}  & \underline{60.9}  & 57.4  & \underline{62.8}  \\
    \textbf{Ours} & \textbf{65.9}  & \underline{69.8}  & \textbf{63.8} & \textbf{66.5} & \textbf{64.8} & \textbf{60.2} & \underline{66.5}  & \textbf{58.8} & \textbf{61.8} & \textbf{58.5} & \textbf{63.7} \\
    \bottomrule
    \end{tabular}
  \label{Scomp1}
\end{table*}

\begin{table*}[htbp]
  \centering
  \caption{\textbf{Accuracy of the generated AVVP pseudo-labels. }The results in the upper part of the table are measured on the validation set of LLP, while the results in the lower part are measured on the test set. The best results are bolded, while the second-best results are underlined.}
  \begin{tabular}{c|c|ccccc|ccccc|c}
  \toprule
  \multirow{2.5}{*}{Set} & \multirow{2.5}{*}{Method} & \multicolumn{5}{c|}{Segment-Level} & \multicolumn{5}{c|}{Event-Level} & \multirow{2.5}{*}{Avg.} \\
  \cmidrule{3-12}
    & & A & V & AV & Type & Event & A & V & AV & Type & Event & \\
  \midrule
  \multirow{3}{*}{Val} & VALOR \cite{lai2023modality} & \underline{80.8} & 72.3 & 65.2 & 72.8 & 79.5 & 71.7 & 66.4 & 57.5 & 65.2 & 67.7 & 69.9 \\
    & UWAV \cite{lai2025uwav} & 80.1 & \textbf{75.4} & \underline{68.5} & \underline{74.7} & \underline{79.9} & \underline{72.1} & \underline{70.7} & \underline{61.4} & \underline{68.1} & \underline{70.5} & \underline{72.1} \\
    & \textbf{EAR (Ours)} & \textbf{83.4} & \underline{75.3} & \textbf{70.3} & \textbf{76.3} & \textbf{81.5} & \textbf{76.9} & \textbf{70.8} & \textbf{65.2} & \textbf{71.0} & \textbf{73.8} & \textbf{74.5} \\
  \midrule
  \multirow{6}{*}{Test} & HAN+PPL \cite{rachavarapu2024weakly} & 62.5 & 55.3 & 52.3 & 56.0 & 58.3 & 55.4 & 51.1 & 46.9 & 50.9 & 50.6 & 53.9 \\
    & MA+PPL \cite{rachavarapu2024weakly} & 61.7 & 61.8 & 57.5 & 60.6 & 59.4 & 55.4 & 57.9 & 51.6 & 55.0 & 52.6 & 57.4 \\
    & VAPLAN \cite{zhou2024advancing} & \underline{80.4} & 73.0 & -- & -- & -- & 71.7 & 68.3 & -- & -- & -- & -- \\
    & VALOR \cite{lai2023modality} & \underline{80.4} & 71.8 & 63.7 & 72.0 & \underline{79.7} & \underline{72.2} & 65.9 & 55.6 & 64.6 & 68.0 & 69.4 \\
    & UWAV \cite{lai2025uwav} & 78.4 & \underline{74.5} & \underline{65.5} & \underline{72.8} & 78.4 & 71.1 & \underline{69.6} & \underline{57.7} & \underline{66.1} & \underline{69.0} & \underline{70.3} \\
    & \textbf{EAR (Ours)} & \textbf{81.7} & \textbf{75.5} & \textbf{68.0} & \textbf{75.1} & \textbf{80.8} & \textbf{75.1} & \textbf{71.3} & \textbf{62.0} & \textbf{69.5} & \textbf{72.9} & \textbf{73.2} \\
  \bottomrule
  \end{tabular}
  \label{Spseudo}
\end{table*}

\subsection{Loss Functions}
The loss function in the main text consists of three components: segment-level supervision with class-imbalance weighting $(\mathcal{L}_{soft}^{A}, \mathcal{L}_{soft}^{V})$, self-supervised regularization terms $(\mathcal{L}_{mix}^{A}, \mathcal{L}_{mix}^{V})$, and video-level supervision $(\mathcal{L}_{video})$. We provide the complete formulation of each part as follows: 

\noindent \textbf{Class-Balanced Supervision with Uncertainty:}	
Following \cite{lai2025uwav}, we also adopt soft pseudo-labels to evaluate the confidence scores. The confidence scores serve as a measure of the generator's uncertainty about the predicted pseudo-labels. The soft pseudo-labels $\hat{P}_{m}$ are shown in Eq. (9) in the main text. Then, the outputs of the classifier $\{P_{A},P_{V}\}$ are constrained by $\hat{P}_{m}$ using the binary cross-entropy loss. However, since event occurrences for most categories are sparse (negative events), with only a few event categories frequently appearing in videos (positive events), it is necessary to weight the losses of different categories to mitigate the adverse effects of class imbalance: 
\begin{equation}
\begin{split}
\mathcal{L}_{soft}^{m} = & w_{pos}^{m} \cdot y \cdot \text{BCE}(P_{m}, \hat{P}_{m}) \\
& + w_{neg}^{m} \cdot (1-y) \cdot \text{BCE}(P_{m}, \hat{P}_{m})
\end{split}
\end{equation}
\begin{equation}
w_{pos}^{m} = \frac{\sum_{i=1}^{N}\sum_{t=1}^{T}\sum_{c=1}^{C} (1 - \hat{y}_{i,t,c}^{m})}{NTC} \times W
\end{equation}
\begin{equation}
w_{neg}^{m} = \frac{\sum_{i=1}^{N}\sum_{t=1}^{T}\sum_{c=1}^{C} \hat{y}_{i,t,c}^{m}}{NTC}
\end{equation}
where $m\in\{A,V\}$, $N$ is the number of videos in the training set, $W$ is a hyper-parameter.

\noindent \textbf{Uncertainty-weighted Feature Mixup:}	
To further enhance the generalization ability of the model, we adopt the self-supervised regularization as in \cite{lai2025uwav}. By mixing the features and pseudo-labels of any two segments, the model is guided to perform prediction on mixed segments: 
\begin{equation}
\bar{f}_{Ae}^{i,j} = \gamma{f}_{Ae}^{i} + (1-\gamma){f}_{Ae}^{j},\quad
\bar{p}_{A}^{i,j} = \gamma\hat{p}_{A}^{i} + (1-\gamma)\hat{p}_{A}^{j}
\end{equation}
\begin{equation}
\bar{f}_{Ve}^{i,j} = \gamma{f}_{Ve}^{i} + (1-\gamma){f}_{Ve}^{j},\quad
\bar{p}_{V}^{i,j} = \gamma\hat{p}_{V}^{i} + (1-\gamma)\hat{p}_{V}^{j}
\end{equation}
where $i$ and $j$ are segment indices, ${f}_{me}^{i}\in F_{me}$, $\hat{p}_{m}^{i}\in \hat{P}_{m}$, $\bar{f}_{me}\in \bar{F}_{me}$, $\bar{p}_{m}\in \bar{P}_{m}$, $m\in\{A,V\}$, $\gamma \sim \operatorname{Beta}(\alpha, \alpha)$ and $\alpha$ is a hyper-parameter of the Beta distribution. Afterward, the mixed uni-modal features $\bar{F}_{me}$ pass through the classifier and a sigmoid activation layer to produce $\bar{P}_{m}^{mix}$, which are constrained by $\bar{P}_{m}$ to participate in self-supervised training:
\begin{equation}
\mathcal{L}_{mix}^{m} = \text{BCE}(\bar{P}_{m}^{mix}, \bar{P}_{m})
\end{equation}

\noindent \textbf{Video-Level Supervision:} 
To provide comprehensive semantic guidance to the model and reduce the impact of noise from pseudo-labels, we employ coarse-grained labels $Y$ for global supervision. The attentive multi-modal multiple instance learning (MMIL) pooling \cite{tian2020unified} is applied to aggregate information in $\{P_{A}, P_{V}\}$ across the temporal and modal dimensions, yielding video-level event probabilities $P$. The video-level loss is:
\begin{equation}
\label{Spreloss}
\mathcal{L}_{video} = \text{BCE}(P, Y)
\end{equation}

\begin{table*}[htbp]
  \centering
  \caption{\textbf{Ablation study on the pseudo-label generator. }"LM" denotes our similarity-based uni-modal label migration. The best results are bolded.}
    \begin{tabular}{c|ccccc|ccccc|c}
    \toprule
    \multirow{2.5}{*}{Method} & \multicolumn{5}{c|}{Segment-Level}    & \multicolumn{5}{c|}{Event-Level}      & \multirow{2.5}{*}{Avg.} \\
\cmidrule{2-11}          & A     & V     & AV    & Type  & Event & A     & V     & AV    & Type  & Event &  \\
    \midrule
    VALOR & 80.4  & 71.8  & 63.7  & 72.0  & 79.7  & 72.2  & 65.9  & 55.6  & 64.6  & 68.0  & 69.4  \\
    UWAV  & 78.4  & 74.5  & 65.5  & 72.8  & 78.4  & 71.1  & 69.6  & 57.7  & 66.1  & 69.0  & 70.3  \\
    UWAV+LM & 79.8  & 73.8  & 65.0  & 72.9  & 79.4  & 72.9  & 69.0  & 58.0  & 66.6  & 70.8  & 70.8  \\
    EAR w/o LM & 81.1  & 74.3  & 66.5  & 74.0  & 80.2  & 74.7  & 69.7  & 60.5  & 68.3  & 71.8  & 72.1  \\
    \textbf{EAR(Ours)} & \textbf{81.7} & \textbf{75.5} & \textbf{68.0} & \textbf{75.1} & \textbf{80.8} & \textbf{75.1} & \textbf{71.3} & \textbf{62.0} & \textbf{69.5} & \textbf{72.9} & \textbf{73.2} \\
    \bottomrule
    \end{tabular}
  \label{Spseudo_structure}
\end{table*}

\begin{table*}[htbp]
  \centering
  \caption{\textbf{Ablation study on the loss functions during pre-training. }$\{\mathcal{L}_{A}, \mathcal{L}_{V}\}$ and $\{\mathcal{L}_{AV_{1}}, \mathcal{L}_{AV_{2}}\}$ denote uni-modal are multi-modal supervision. The best results are bolded.}
    \begin{tabular}{cccc|ccccc|ccccc|c}
    \toprule
    \multirow{2.5}{*}{$\mathcal{L}_{A}$} & \multirow{2.5}{*}{$\mathcal{L}_{V}$} & \multirow{2.5}{*}{$\mathcal{L}_{AV_{1}}$} & \multirow{2.5}{*}{$\mathcal{L}_{AV_{2}}$} & \multicolumn{5}{c|}{Segment-Level}    & \multicolumn{5}{c|}{Event-Level}      & \multirow{2.5}{*}{Avg.} \\
\cmidrule{5-14}          &       &       &       & A     & V     & AV    & Type  & Event & A     & V     & AV    & Type  & Event &  \\
    \midrule
    $\times$     & $\checkmark$     & $\checkmark$     & $\checkmark$     & 81.2  & 74.9  & 67.5  & 74.6  & 80.4  & 74.8  & 70.4  & \textbf{62.1} & 69.1  & 72.2  & 72.7  \\
    $\checkmark$     & $\times$     & $\checkmark$     & $\checkmark$     & 80.3  & 74.3  & 66.2  & 73.6  & 79.5  & 73.3  & 69.9  & 60.0  & 67.7  & 70.8  & 71.6  \\
    $\checkmark$     & $\checkmark$     & $\times$     & $\checkmark$     & 80.1  & 74.2  & 65.9  & 73.4  & 79.5  & 73.2  & 69.7  & 59.5  & 67.5  & 71.0  & 71.4  \\
    $\checkmark$     & $\checkmark$     & $\checkmark$     & $\times$     & 78.6  & 72.4  & 63.3  & 71.4  & 78.3  & 71.0  & 68.0  & 56.5  & 65.1  & 68.8  & 69.3  \\
    $\times$     & $\times$     & $\checkmark$     & $\checkmark$     & 81.1  & 74.3  & 66.5  & 74.0  & 80.2  & 74.7  & 69.7  & 60.5  & 68.3  & 71.8  & 72.1  \\
    $\checkmark$     & $\checkmark$     & $\checkmark$     & $\checkmark$     & \textbf{81.7} & \textbf{75.5} & \textbf{68.0} & \textbf{75.1} & \textbf{80.8} & \textbf{75.1} & \textbf{71.3} & 62.0  & \textbf{69.5} & \textbf{72.9} & \textbf{73.2} \\
    \bottomrule
    \end{tabular}
  \label{Spseudo_loss}
\end{table*}

\begin{table*}[htbp]
  \centering
  \caption{\textbf{Ablation studies on pseudo-label performance and model performance. }The best results are bolded, while the second-best results are underlined.}
    \begin{tabular}{c|ccccc|ccccc|c}
    \toprule
    \multirow{2.5}{*}{Method} & \multicolumn{5}{c|}{Segment-Level}    & \multicolumn{5}{c|}{Event-Level}      & \multirow{2.5}{*}{Avg.} \\
\cmidrule{2-11}          & A     & V     & AV    & Type  & Event & A     & V     & AV    & Type  & Event &  \\
    \midrule
    HAN(VALOR) & 62.8  & 66.7  & 60.0  & 63.2  & 62.3  & 57.1  & 63.9  & 54.4  & 58.5  & 55.9  & 60.5  \\
    HAN(UWAV) & 64.2  & \textbf{70.0} & \underline{63.4}  & \underline{65.9}  & 63.9  & 58.6  & \textbf{66.7} & 57.5  & 60.9  & 57.4  & 62.8  \\
    HAN(EAR) & \underline{65.7}  & 68.7  & 63.1  & 65.8  & \underline{64.5}  & \textbf{60.4} & 65.4  & \underline{58.2}  & \underline{61.3}  & \underline{58.3}  & \underline{63.1}  \\
    \textbf{ERM(Ours)} & \textbf{65.8} & \underline{69.8}  & \textbf{63.8} & \textbf{66.5} & \textbf{64.8} & \underline{60.2}  & \underline{66.5}  & \textbf{58.8} & \textbf{61.8} & \textbf{58.5} & \textbf{63.7} \\
    \bottomrule
    \end{tabular}
  \label{Spseudo_han}
\end{table*}

\begin{table*}[htbp]
  \centering
  \caption{\textbf{Ablation study on the AVVP model with soft constraints. }"AMDF" denotes the Asymmetric Audio/Visual-Driven Fusion module, and "ERM" denotes the Multi-Event Relationship Modeling module. The best results are bolded.}
    \begin{tabular}{cc|rrrrr|rrrrr|r}
    \toprule
    \multirow{2.5}{*}{AMDF} & \multirow{2.5}{*}{ERM} & \multicolumn{5}{c|}{Segment-Level}    & \multicolumn{5}{c|}{Event-Level}      & \multicolumn{1}{c}{\multirow{2.5}{*}{Avg.}} \\
\cmidrule{3-12}          &       & \multicolumn{1}{c}{A} & \multicolumn{1}{c}{V} & \multicolumn{1}{c}{AV} & \multicolumn{1}{c}{Type} & \multicolumn{1}{c|}{Event} & \multicolumn{1}{c}{A} & \multicolumn{1}{c}{V} & \multicolumn{1}{c}{AV} & \multicolumn{1}{c}{Type} & \multicolumn{1}{c|}{Event} &  \\
    \midrule
    $\times$     & $\checkmark$     & 64.9  & 68.4  & 62.3  & 65.2  & 64.0  & 59.4  & 64.8  & 56.6  & 60.3  & 57.4  & 62.3  \\
    $\checkmark$     & $\times$     & 65.1  & 68.9  & 63.2  & 65.7  & 64.4  & 59.2  & 65.4  & 57.7  & 60.8  & 57.6  & 62.8  \\
    $\checkmark$     & $\checkmark$     & \textbf{65.8} & \textbf{69.8} & \textbf{63.8} & \textbf{66.5} & \textbf{64.8} & \textbf{60.2} & \textbf{66.5} & \textbf{58.8} & \textbf{61.8} & \textbf{58.5} & \textbf{63.7} \\
    \bottomrule
    \end{tabular}
  \label{Savvp_structure}
\end{table*}

\begin{table*}[htbp]
  \centering
  \caption{\textbf{Ablation study of different structure in the Asymmetric Audio/Visual-Driven Fusion module. } The best results are bolded.}
    \begin{tabular}{c|ccccc|ccccc|c}
    \toprule
    \multirow{2.5}{*}{Method} & \multicolumn{5}{c|}{Segment-Level}    & \multicolumn{5}{c|}{Event-Level}      & \multirow{2.5}{*}{Avg.} \\
\cmidrule{2-11}          & A     & V     & AV    & Type  & Event & A     & V     & AV    & Type  & Event &  \\
    \midrule
    MSA+MCA & 65.1  & 68.5  & 62.9  & 65.5  & 64.0  & 59.4  & 65.4  & 57.4  & 60.7  & 57.6  & 62.7  \\
    HAN   & 65.1  & 68.9  & 63.0  & 65.7  & 64.4  & 59.7  & 65.8  & 58.0  & 61.2  & 58.0  & 63.0  \\
    \textbf{AMDF(Ours)} & \textbf{65.8} & \textbf{69.8} & \textbf{63.8} & \textbf{66.5} & \textbf{64.8} & \textbf{60.2} & \textbf{66.5} & \textbf{58.8} & \textbf{61.8} & \textbf{58.5} & \textbf{63.7} \\
    \bottomrule
    \end{tabular}
  \label{Savvp_amdf}
\end{table*}

\begin{table*}[htbp]
  \centering
  \caption{\textbf{Ablation study of different structure in the Multi-Event Relationship Modeling module. } The best results are bolded, while the second-best results are underlined.}
    \begin{tabular}{c|ccccc|ccccc|c}
    \toprule
    \multirow{2.5}{*}{Method} & \multicolumn{5}{c|}{Segment-Level}    & \multicolumn{5}{c|}{Event-Level}      & \multirow{2.5}{*}{Avg.} \\
\cmidrule{2-11}          & A     & V     & AV    & Type  & Event & A     & V     & AV    & Type  & Event &  \\
    \midrule
    Attention & 64.5  & 67.7  & 62.0  & 64.7  & 63.6  & 59.3  & 64.6  & 57.1  & 60.3  & 57.3  & 62.1  \\
    ERM A/V-AV & \textbf{66.2} & \underline{69.2}  & \underline{63.6}  & \underline{66.3}  & \textbf{65.2} & 60.0  & \underline{66.0}  & 58.2  & \underline{61.4}  & 58.3  & \underline{63.4}  \\
    ERM w/o A/V & 65.2  & 68.6  & 63.1  & 65.7  & 64.3  & 59.4  & 65.3  & 57.5  & 60.7  & 57.7  & 62.8  \\
    ERM w/o AV & 65.5  & 68.4  & 63.0  & 65.6  & 64.6  & \textbf{60.5} & 65.1  & \underline{58.4}  & 61.3  & \textbf{58.7} & 63.1  \\
    \textbf{ERM(Ours)} & \underline{65.8}  & \textbf{69.8} & \textbf{63.8} & \textbf{66.5} & \underline{64.8}  & \underline{60.2}  & \textbf{66.5} & \textbf{58.8} & \textbf{61.8} & \underline{58.5}  & \textbf{63.7} \\
    \bottomrule
    \end{tabular}
  \label{Savvp_erm}
\end{table*}

\begin{table*}[htbp]
  \centering
  \caption{\textbf{Ablation study on the similarity thresholds in uni-modal label migration. } The best results are bolded, while the second-best results are underlined.}
    \begin{tabular}{cc|ccccc|ccccc|c}
    \toprule
    \multirow{2.5}{*}{$\mu_{A}$} & \multirow{2.5}{*}{$\mu_{V}$} & \multicolumn{5}{c|}{Segment-Level}    & \multicolumn{5}{c|}{Event-Level}      & \multirow{2.5}{*}{Avg.} \\
\cmidrule{3-12}          &       & A     & V     & AV    & Type  & Event & A     & V     & AV    & Type  & Event &  \\
    \midrule
    0.97  & 0.95  & 80.4  & \textbf{76.0} & 67.3  & 74.6  & 80.2  & 73.3  & \textbf{71.8} & 60.9  & 68.6  & 71.6  & 72.5  \\
    0.99  & 0.95  & 80.0  & \underline{75.5}  & 66.5  & 74.0  & 80.0  & 72.8  & \underline{71.5}  & 60.2  & 68.2  & 71.3  & 72.0  \\
    \midrule
    0.98  & 0.92  & 80.0  & 73.9  & 64.9  & 72.9  & 79.4  & 73.2  & 69.4  & 58.2  & 66.9  & 70.9  & 71.0  \\
    0.98  & 0.94  & \underline{81.4}  & 72.6  & 66.0  & 73.3  & 79.4  & 74.2  & 68.5  & 60.7  & 67.8  & 71.5  & 71.5  \\
    0.98  & 0.96  & \textbf{81.7} & 74.8  & \underline{67.5}  & \underline{74.7}  & \underline{80.6}  & \textbf{75.2} & 70.5  & \underline{61.3}  & \underline{69.0}  & \underline{72.5}  & \underline{72.8}  \\
    0.98  & 0.98  & 79.9  & 72.6  & 64.3  & 72.3  & 79.3  & 72.9  & 67.7  & 57.4  & 66.0  & 70.0  & 70.2  \\
    \midrule
    \textbf{0.98} & \textbf{0.95} & \textbf{81.7} & 75.5  & \textbf{68.0} & \textbf{75.1} & \textbf{80.8} & \underline{75.1}  & 71.3  & \textbf{62.0} & \textbf{69.5} & \textbf{72.9} & \textbf{73.2} \\
    \bottomrule
    \end{tabular}
  \label{Spseudo_mu}
\end{table*}

\begin{table*}[htbp]
  \centering
  \caption{\textbf{Ablation study on the weights of uni-modal supervision during the pre-training of the pseudo-label generator. } The best results are bolded.}
    \begin{tabular}{cc|ccccc|ccccc|c}
    \toprule
    \multirow{2.5}{*}{$\lambda_{A}$} & \multirow{2.5}{*}{$\lambda_{V}$} & \multicolumn{5}{c|}{Segment-Level}    & \multicolumn{5}{c|}{Event-Level}      & \multirow{2.5}{*}{Avg.} \\
\cmidrule{3-12}          &       & A     & V     & AV    & Type  & Event & A     & V     & AV    & Type  & Event &  \\
    \midrule
    0     & 0.15  & \underline{81.2}  & 74.9  & \underline{67.5}  & \underline{74.6}  & \underline{80.4}  & \underline{74.8}  & 70.4  & \textbf{62.1} & \underline{69.1}  & \underline{72.2}  & \underline{72.7}  \\
    0.1   & 0.15  & 79.9  & 75.4  & 66.2  & 73.9  & 79.9  & 72.7  & 71.2  & 59.3  & 67.7  & 71.1  & 71.7  \\
    0.15  & 0.15  & 79.6  & 74.2  & 65.0  & 72.9  & 79.4  & 72.5  & 69.5  & 58.1  & 66.7  & 70.6  & 70.8  \\
    \midrule
    0.05  & 0.05  & 80.0  & \textbf{76.1} & 66.8  & 74.3  & 80.1  & 72.9  & \textbf{71.6} & 60.3  & 68.3  & 71.2  & 72.2  \\
    0.05  & 0.25  & 79.8  & 73.4  & 65.4  & 72.9  & 79.2  & 73.0  & 69.0  & 59.5  & 67.2  & 70.5  & 71.0  \\
    \midrule
    \textbf{0.05} & \textbf{0.15} & \textbf{81.7} & \underline{75.5}  & \textbf{68.0} & \textbf{75.1} & \textbf{80.8} & \textbf{75.1} & \underline{71.3}  & \underline{62.0}  & \textbf{69.5} & \textbf{72.9} & \textbf{73.2} \\
    \bottomrule
    \end{tabular}
  \label{Spseudo_lambda}
\end{table*}

\begin{table*}[htbp]
  \centering
  \caption{\textbf{Ablation study on the layers of the Multi-Event Relationship Modeling. } The best results are bolded.}
    \begin{tabular}{c|rrrrr|rrrrr|r}
    \toprule
    \multirow{2.5}{*}{$M$} & \multicolumn{5}{c|}{Segment-Level}    & \multicolumn{5}{c|}{Event-Level}      & \multicolumn{1}{c}{\multirow{2.5}{*}{Avg.}} \\
\cmidrule{2-11}          & \multicolumn{1}{c}{A} & \multicolumn{1}{c}{V} & \multicolumn{1}{c}{AV} & \multicolumn{1}{c}{Type} & \multicolumn{1}{c|}{Event} & \multicolumn{1}{c}{A} & \multicolumn{1}{c}{V} & \multicolumn{1}{c}{AV} & \multicolumn{1}{c}{Type} & \multicolumn{1}{c|}{Event} &  \\
    \midrule
    2     & 65.8  & 69.3  & 63.4  & 66.2  & \textbf{64.8} & 60.4  & 65.8  & 58.3  & 61.5  & 58.4  & 63.4  \\
    \textbf{3} & 65.8  & \textbf{69.8} & \textbf{63.8} & \textbf{66.5} & \textbf{64.8} & 60.2  & \textbf{66.5} & \textbf{58.8} & \textbf{61.8} & \textbf{58.5} & \textbf{63.7} \\
    4     & 65.8  & 68.1  & 62.5  & 65.5  & 64.4  & \textbf{60.8} & 64.9  & 57.2  & 61.0  & 58.6  & 62.9  \\
    \bottomrule
    \end{tabular}
  \label{Savvp_m}
\end{table*}

\section{Experimental Details}
\subsection{Comparative Studies}
In this subsection, we provide complete comparative experiments, including the audio-visual video parsing performance and pseudo-label performance of different methods.

\noindent \textbf{Comparison of AVVP Performance:}	
As shown in Table \ref{Scomp1}, we present a comprehensive comparison with almost all existing AVVP methods using features extracted by VGGish+ResNet \cite{hershey2017cnn, he2016deep, tran2018closer}. Compared to existing methods, EAR achieves the best or second-best performance across all segment-level and event-level metrics, attaining the highest average video parsing performance of 63.7\%. In both detection and localization of uni-modal and multi-modal events, our method sets a new state-of-the-art benchmark. These results demonstrate the effectiveness and generalization of our proposed framework.

\noindent \textbf{Accuracy of Generated Pseudo-Labels:}
For the pseudo-labels produced by our generator for the target dataset, we evaluate their accuracy on the validation the test sets, with the results shown in Table \ref{Spseudo}. Among these, we only report results that are verifiable either from the papers or through available code. On the validation set of LLP \cite{tian2020unified}, EAR achieves the best or second-best performance across all metrics, while on the test set, it outperforms all existing state-of-the-art methods on all metric. The average AVVP performance of our generated pseudo-labels reaches a new benchmark, surpassing the current state-of-the-art methods VALOR \cite{lai2023modality} and UWAV \cite{lai2025uwav} by 4.6\% and 2.4\% on the validation set, and exceeding them by 3.8\% and 2.9\% on the test set, respectively. In summary, the proposed pseudo-label generator can achieve enhanced perception of uni-modal and multi-modal events, thus generating more accurate AVVP pseudo-labels. We will release these pseudo-labels to provide fine-grained guidance for future research. 

\subsection{Ablation Studies}
In this subsection, we provide complete evaluation results for all ablation studies in the main text, as shown in Tables \ref{Spseudo_structure} to \ref{Savvp_erm}. In addition, we perform parameter ablation studies, and the results are shown in Tables \ref{Spseudo_mu} to \ref{Savvp_m}. In this supplementary material, we present analyses only for the parameter ablation studies, while analyses of the other ablation studies are included in the main text. 

\noindent \textbf{Impact of the Label Migration Thresholds:}
In Table \ref{Spseudo_mu}, we show how the similarity thresholds $\mu_{A}$ and $\mu_{V}$ for uni-modal label migration affect the performance of the pseudo-label generator. With $\mu_{A}=0.95$ and $\mu_{V}=0.98$, the generated pseudo-labels achieve the best or second-best performance across most metrics. When $\mu_{A}$ or $\mu_{V}$ are increased or decreased even slightly, the performance of the generated pseudo-labels on the target dataset shows a measurable decline. Larger changes in the similarity thresholds $\mu_{V}$ have a more pronounced impact on the performance of the pseudo-label generator. Moreover, increasing $\mu_{A}$ and $\mu_{V}$ has a more negative effect on the performance of the pseudo-label generator than decreasing them. Because the uni-modal labels migrated under low thresholds are relatively complete (despite some noise), while those under high thresholds have a high rate of missed annotations. Additionally, we observe an interesting phenomenon: using an inappropriate $\mu_{A}$ reduces the audio parsing capability of the model, while conversely enhancing the visual parsing capability, and vice versa. We speculate that noise in the supervision of one modality compels the model to focus on optimizing the parsing of the other modality. To sum up, the migration of labels from audio-visual events to uni-modal events is sensitive to the similarity thresholds, which must be carefully adjusted to preserve correctness of uni-modal labels.

\noindent \textbf{Impact of the Weights of Uni-Modal Supervision:}
As shown in Table \ref{Spseudo_lambda}, we study the impact of the weights for uni-modal supervision on the performance of the pseudo-label generator. With $\lambda_{A}=0.05$ and $\lambda_{V}=0.15$, our generator achieves the best performance across all metrics. Increasing the weights of the uni-modal supervision may amplify the influence of noise in the migrated labels, thereby reducing the effectiveness of pre-training for the generator. In contrast, reducing $\lambda_{A}$ or $\lambda_{V}$ would weaken the ability of the generator to perceive uni-modal events. Therefore, selecting appropriate weights for the uni-modal loss functions requires a trade-off between semantic guidance from uni-modal events and interference from label noise, ensuring proper constraint on the pseudo-label generator during pre-training. 

\noindent \textbf{Impact of the Layers in MER:}
As shown in Table \ref{Savvp_m}, we also explore the impact of the layers in our multi-event relationship modeling module on model performance. It can be observed that when the number of layers for event relationship modeling is set to 3, the best performance is achieved on nearly all metrics. Redundant event relationship modeling may lead to excessive fusion of audio-visual information, while insufficient event relationship modeling fails to effectively capture dependencies among multiple events. Selecting an appropriate number of layers makes our soft constraints in this module effective, enabling high-level semantic relationship modeling while avoiding excessive loss of uni-modal information. 

\subsection{Qualitative Results}
\begin{figure}[h]
	\centering
	\includegraphics[width=9cm,height=5.27cm]{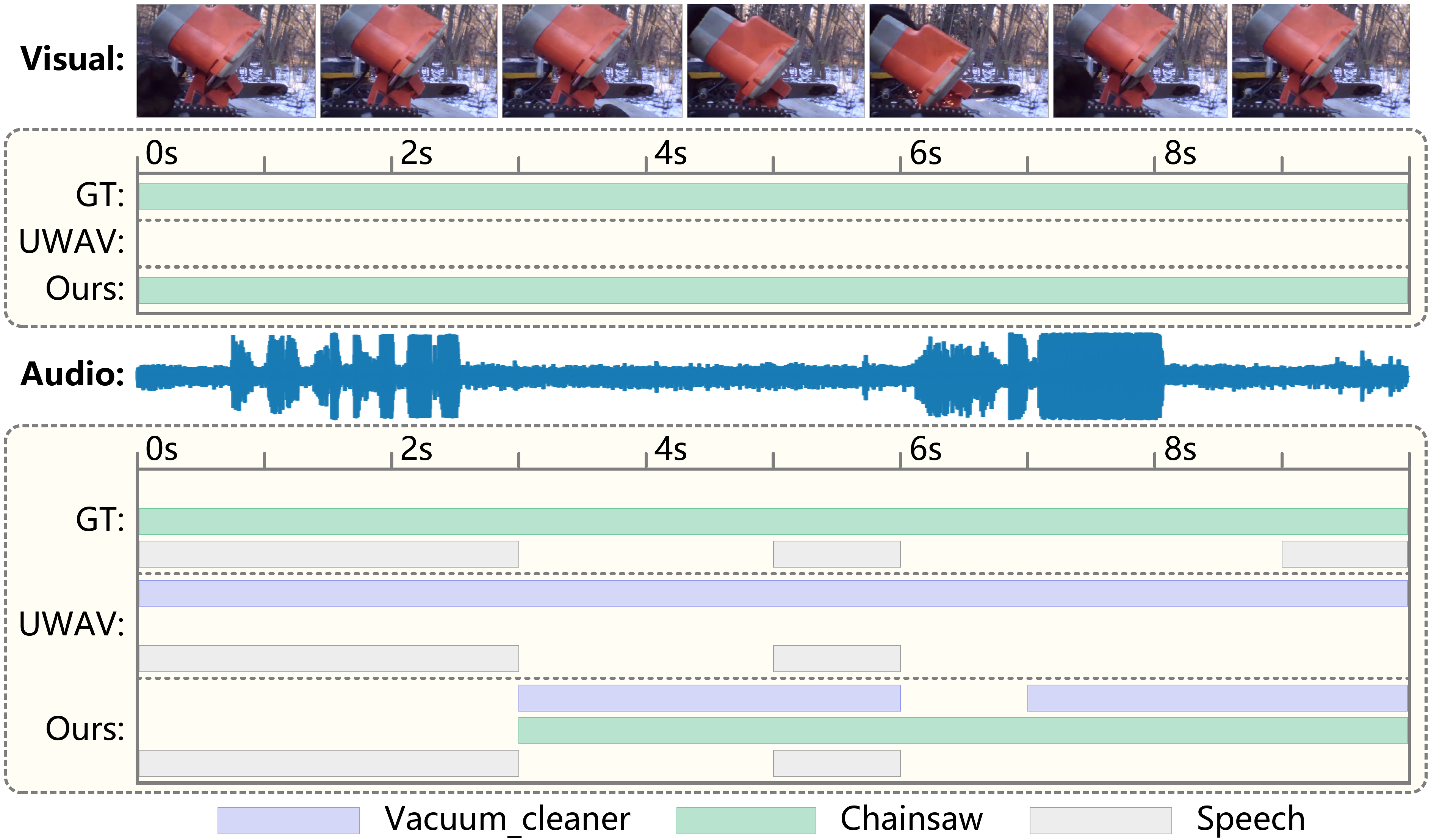}
	\caption{Qualitative Comparison (Example 3).}
	\label{6mWz9fxXO7s}
\end{figure}

\begin{figure}[h]
	\centering
	\includegraphics[width=9cm,height=7.05cm]{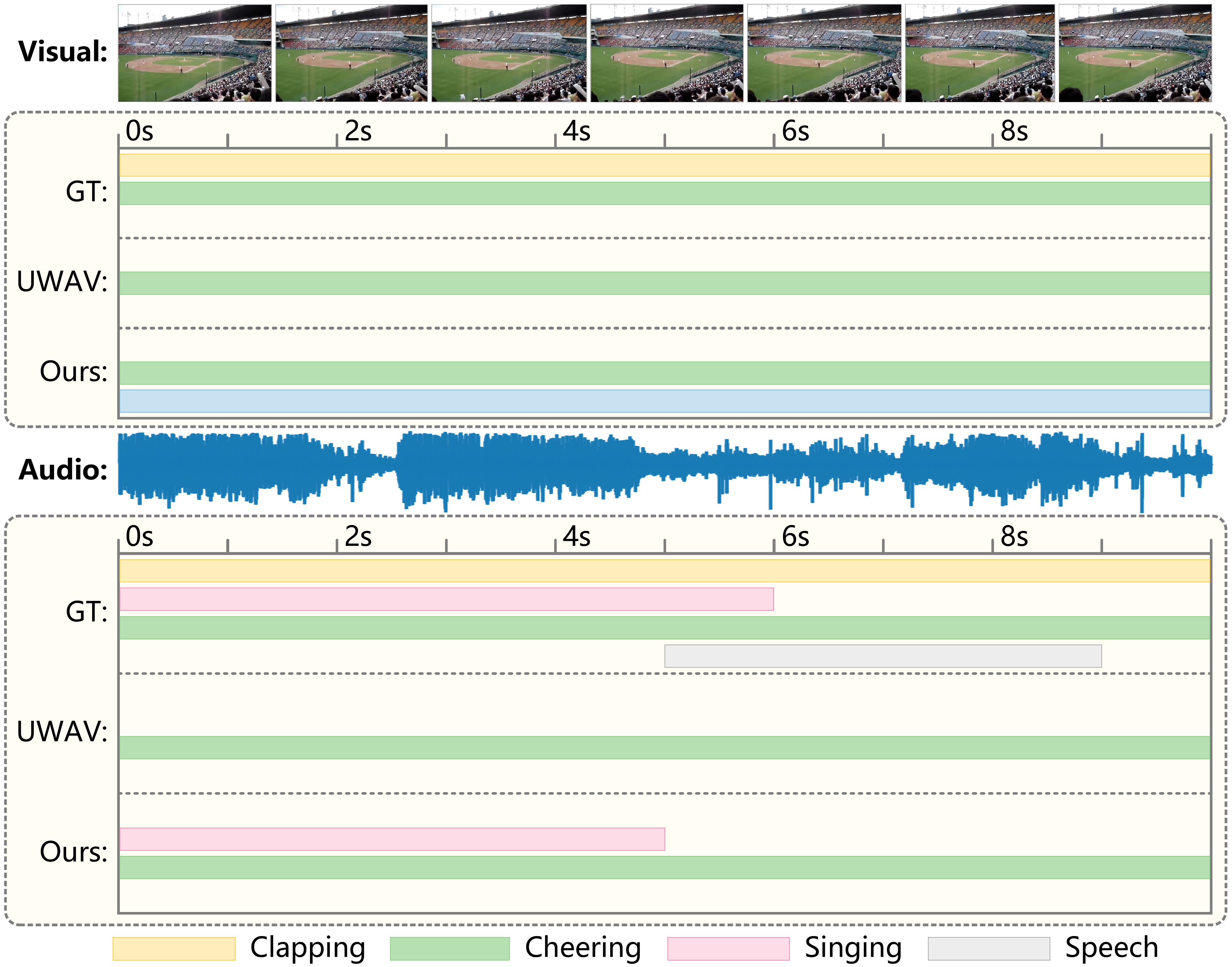}
	\caption{Qualitative Comparison (Example 5).}
	\label{ANNIeZ6xhmM}
\end{figure}

\begin{figure}[h]
	\centering
	\includegraphics[width=9cm,height=5.15cm]{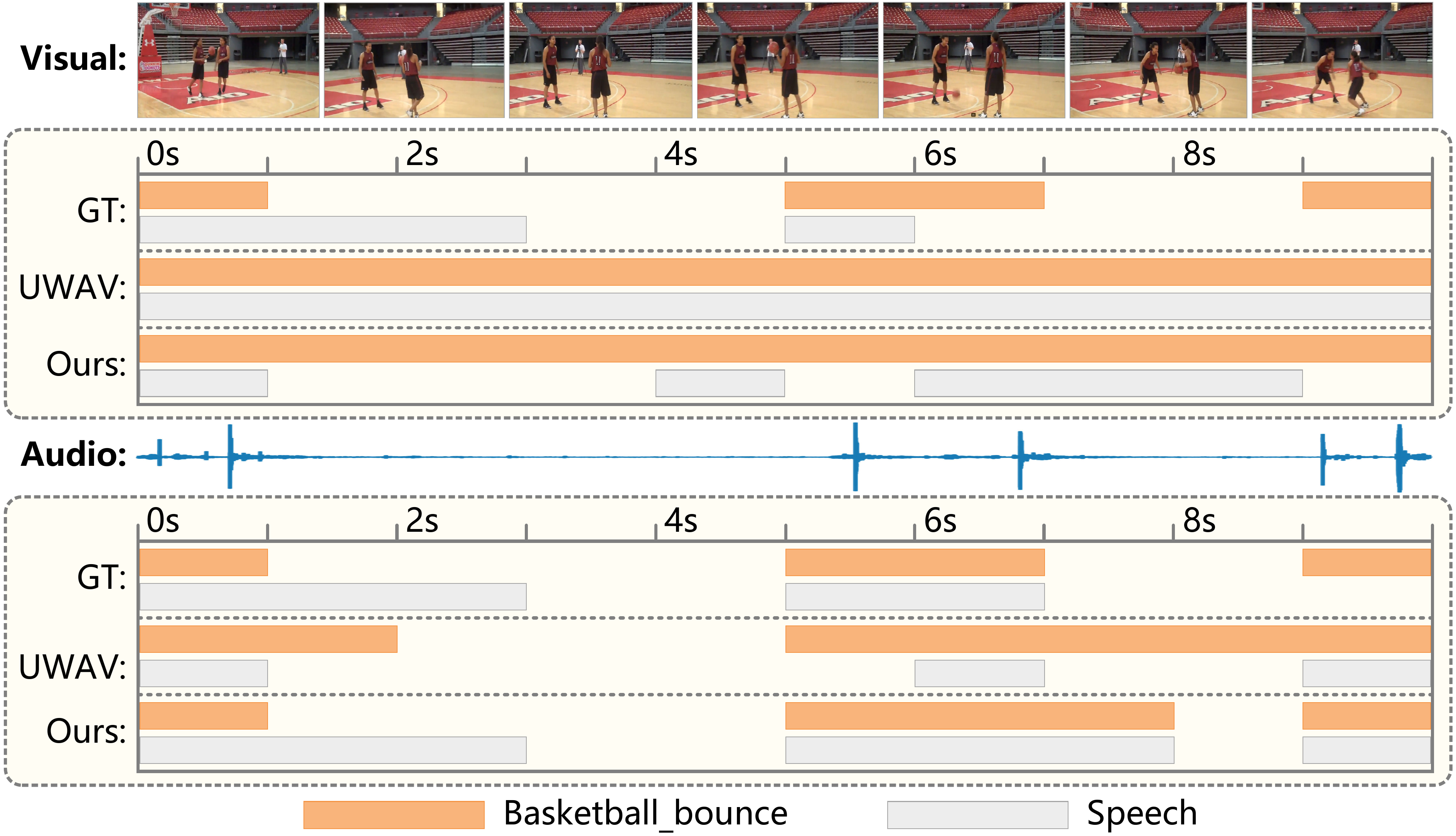}
	\caption{Qualitative Comparison (Example 4).}
	\label{OTattrwc_Qc}
\end{figure}

\begin{figure}[h]
	\centering
	\includegraphics[width=9cm,height=6.37cm]{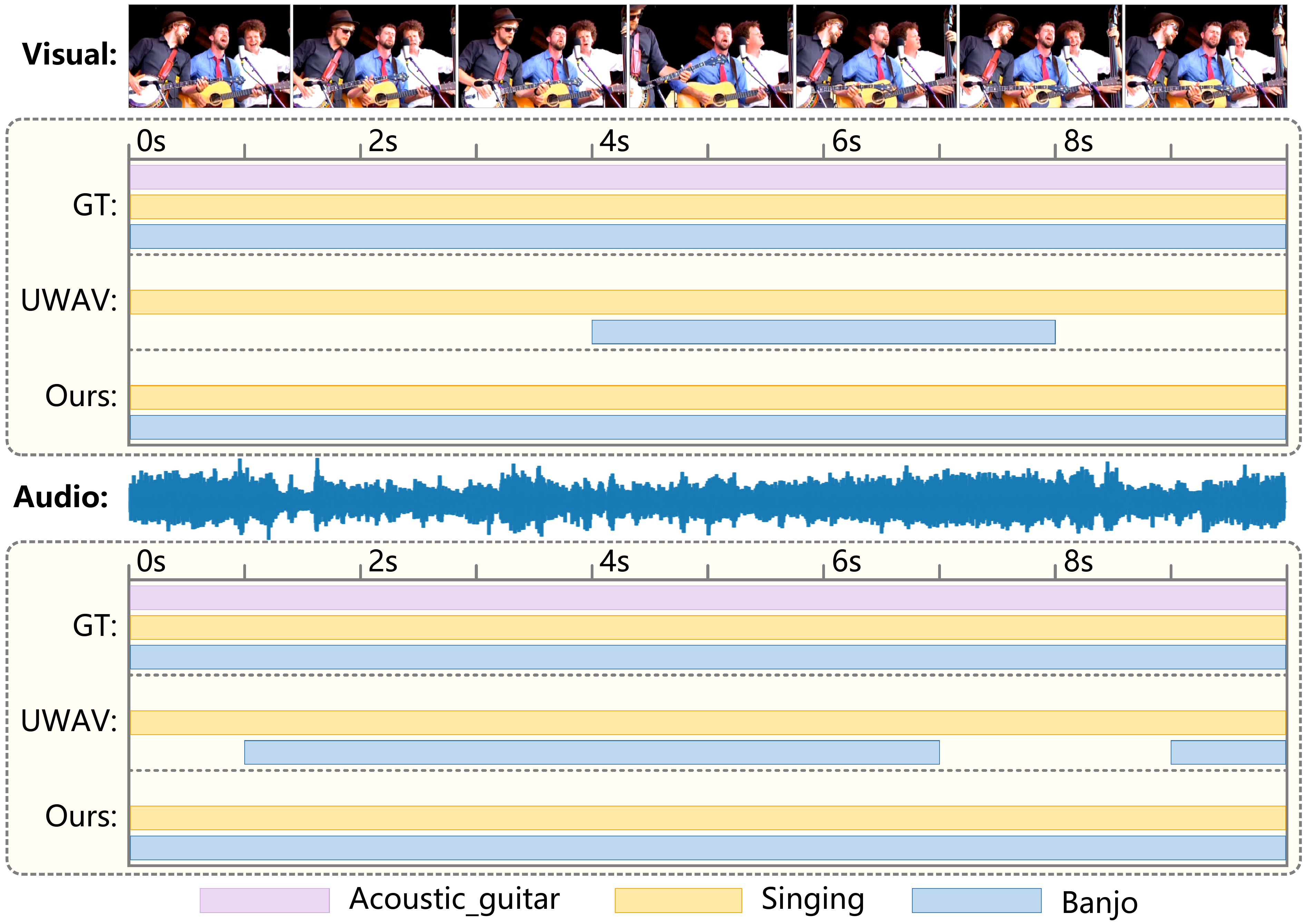}
	\caption{Qualitative Comparison (Example 6).}
	\label{GpPcLJ0b40o}
\end{figure}

\begin{figure}[h]
	\centering
	\includegraphics[width=9cm,height=5.8cm]{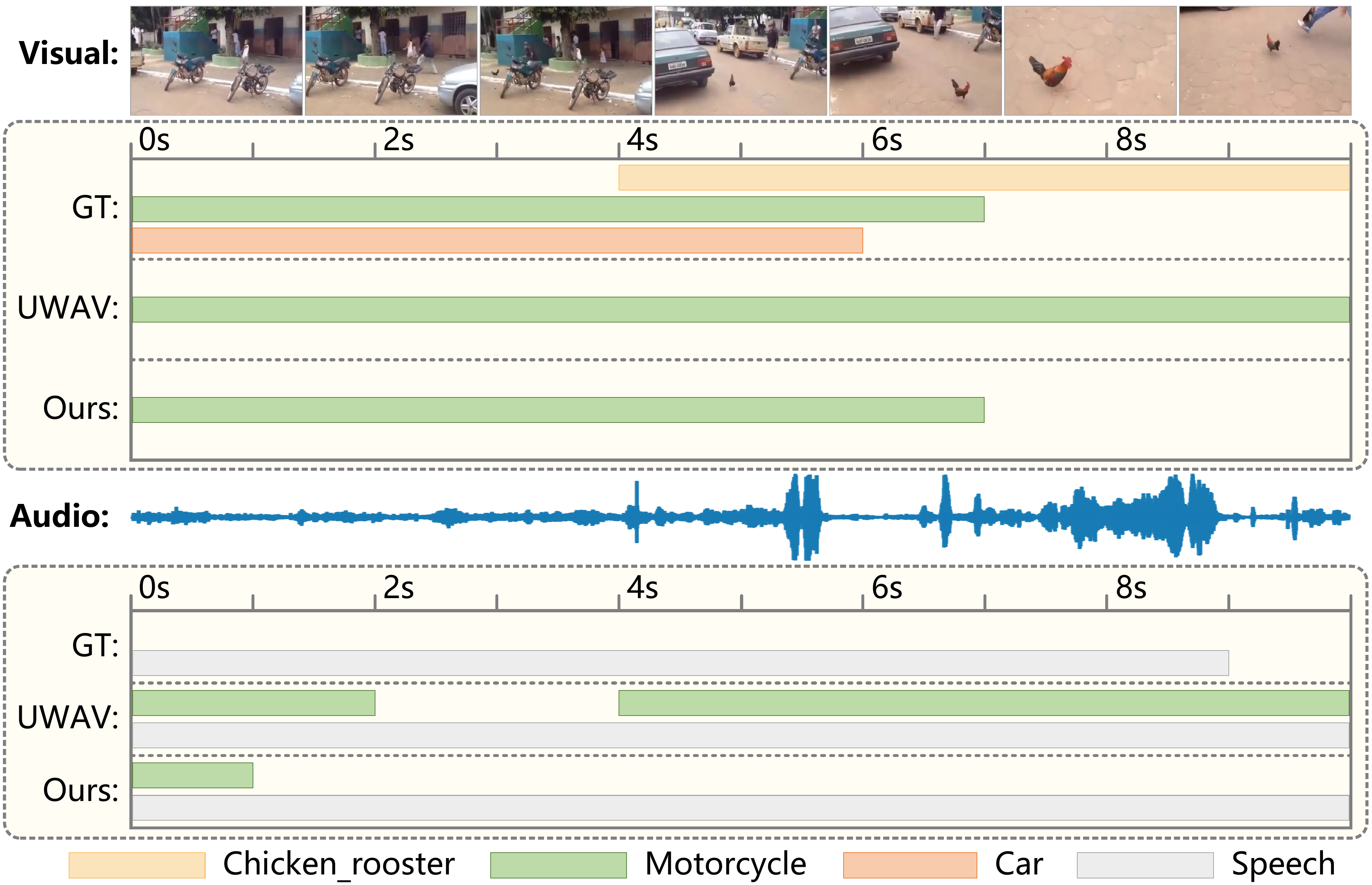}
	\caption{Qualitative Comparison (Example 7).}
	\label{JcunMsOU1g8}
\end{figure}

\begin{figure}[h]
	\centering
	\includegraphics[width=9cm,height=5.17cm]{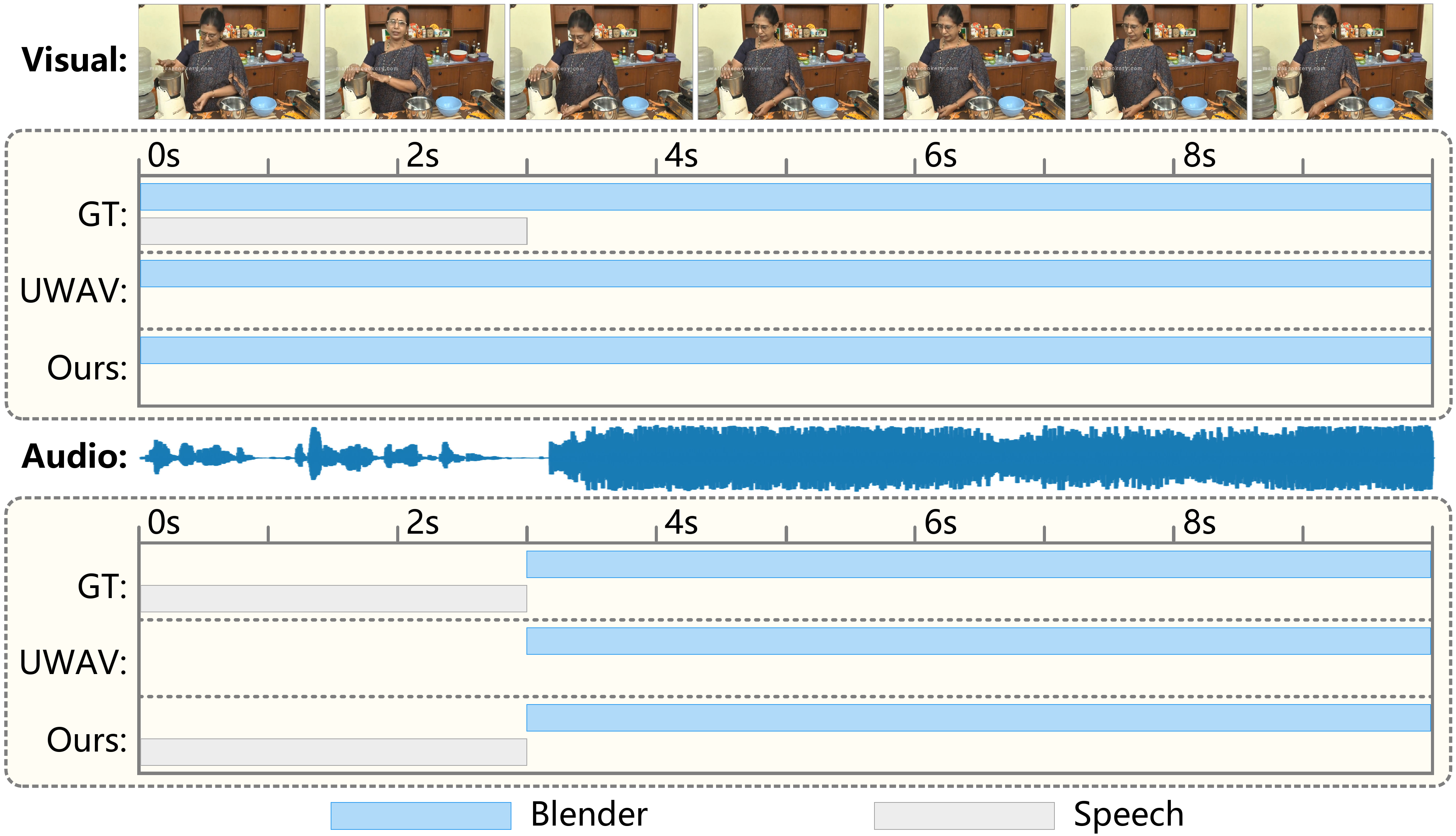}
	\caption{Qualitative Comparison (Example 8).}
	\label{JX-yvRwRrEU}
\end{figure}

In this subsection, we provide additional qualitative comparisons of AVVP to further demonstrate the effectiveness and generalization ability of our EAR. Qualitative results are shown in Figures \ref{6mWz9fxXO7s} to \ref{JX-yvRwRrEU}. From these figures, it can be observed that compared with the state-of-the-art method UWAV \cite{lai2025uwav}, yhe proposed method exhibits fewer instances of missed detection in the videos (e.g., audio and visual events "Chainsaw" in Figure \ref{6mWz9fxXO7s}; audio event "Singing" in Figure \ref{ANNIeZ6xhmM}; audio event "Speech" in Figure \ref{JX-yvRwRrEU}). Moreover, for events that are correctly detected by UWAV \cite{lai2025uwav} and EAR, our method achieves a higher temporal intersection-over-union (tIoU) with the ground truth (e.g., events "Basketball\_bounce" and "Speech" in Figure \ref{OTattrwc_Qc}; audio and visual events "Banjo" in Figure \ref{GpPcLJ0b40o}). Additionally, although EAR sometimes detects events not present in the ground truth, it produces fewer erroneous segments compared to UWAV \cite{lai2025uwav} (e.g., audio event "Vacuum\_cleaner" in Figure \ref{6mWz9fxXO7s}; audio event "Motorcycle" in Figure \ref{JcunMsOU1g8}). Overall, our method achieves accurate temporal perception and semantic classification of audio and visual events, thus attaining superior audio-visual video parsing performance. 
\newpage

\vfill

\end{document}